\documentclass[aps,showpacs,preprintnumbers,superscriptaddress,nofootinbib,floatfix,notitlepage,11pt]{revtex4-1}
\usepackage{amsfonts,amssymb,stmaryrd,latexsym,amsmath,braket}
\usepackage{graphicx,subfigure}
\usepackage{tikz-feynman}
\usepackage{comment}
\usepackage{times}
\usepackage{slashed}
\usepackage{bm}
\usepackage{braket}
\usepackage{mathrsfs} 
\usepackage{xcolor}
\usepackage{xfrac}
\usepackage{hhline}
\usepackage{multirow}
\usepackage{rotating}
\usepackage{setspace} 

\tikzfeynmanset{ 
d3/.style = {
   decoration={
     markings,
     mark=at position 0.5
          with {\arrow[xshift=0mm]{Stealth[black,width=2mm,length=3mm]}\arrow[xshift=2mm]{Stealth[black,width=2mm,length=3mm]}\arrow[xshift=4mm]{Stealth[black,width=2mm,length=3mm]}}
     },
   postaction=decorate},
d2/.style = {
   decoration={
     markings,
     mark=at position 0.5
          with {\arrow[xshift=1mm]{Stealth[black,width=2mm,length=3mm]}\arrow[xshift=3mm]{Stealth[black,width=2mm,length=3mm]}}
     },
   postaction=decorate},
	dn/.style = {
   decoration={
     markings,
     mark=at position 0.5
          with {\arrow[xshift=2mm]{Stealth[black,width=2mm,length=3mm]}}
     },
   postaction=decorate}
}

\makeatletter
\DeclareRobustCommand\bigop[2][1]{%
  \mathop{\vphantom{\sum}\mathpalette\bigop@{{#1}{#2}}}\slimits@
}
\newcommand{\bigop@}[2]{\bigop@@#1#2}
\newcommand{\bigop@@}[3]{%
  \vcenter{%
    \sbox\z@{$#1\sum$}%
    \hbox{\resizebox{\ifx#1\displaystyle#2\fi\dimexpr\ht\z@+\dp\z@}{!}{$\m@th#3$}}%
  }%
}
\makeatother

\newcommand{\mlpsum}{\DOTSB\bigop{\Omega}}

\makeatletter
\let\saved@includegraphics\includegraphics
\AtBeginDocument{\let\includegraphics\saved@includegraphics}
\renewenvironment*{figure}{\@float{figure}}{\end@float}
\makeatother

\begin{document} \singlespacing

\nopagebreak

\bibliographystyle{naturemag}

\title{Neural Network Perturbation Theory and its Application to the Born Series}

\author{Bastian Kaspschak}

\affiliation{Helmholtz-Institut f\"ur Strahlen- und Kernphysik and Bethe
Center
for Theoretical Physics, Universit\"at Bonn, D-53115 Bonn, Germany}

\author{Ulf-G. Mei{\ss}ner}

\affiliation{Helmholtz-Institut f\"ur Strahlen- und Kernphysik and Bethe
Center
for Theoretical Physics, Universit\"at Bonn, D-53115 Bonn, Germany}

\affiliation{Institute for Advanced Simulation, Institut f\"ur Kernphysik,
and
J\"ulich Center for Hadron Physics, Forschungszentrum J\"ulich,
D-52425 J\"ulich, Germany}

\affiliation{Tbilisi State University, 0186 Tbilisi, Georgia}

\begin{abstract}
Deep Learning using the eponymous deep neural networks (DNNs) has become an attractive approach towards 
various data-based problems of theoretical physics in the past decade. There has been a clear trend to
deeper architectures containing increasingly more powerful and involved layers. Contrarily, Taylor 
coefficients of DNNs still appear mainly in the light of interpretability studies, where they are computed
at most to first order. However, especially in theoretical physics numerous problems benefit from 
accessing higher orders, as well.
This gap motivates a general formulation of neural network (NN) Taylor expansions. Restricting our 
analysis to multilayer perceptrons (MLPs) and introducing quantities we refer to as propagators and 
vertices, both depending on the MLP's weights and biases, we establish a graph-theoretical approach.
Similarly to Feynman rules in quantum field theories, we can 
systematically assign diagrams containing propagators and vertices to the corresponding 
partial derivative. 
Examining this approach for S-wave scattering lengths of shallow
potentials, we observe NNs to adapt their derivatives mainly to the leading order of the target
function's Taylor expansion. To circumvent this problem, we propose an iterative NN perturbation
theory. During each iteration we eliminate the leading order, such that the next-to-leading order can
be faithfully learned during the subsequent iteration. After performing two iterations, we find
that the first- and second-order Born terms are correctly adapted during the respective iterations. 
Finally, we combine both results to find a proxy that acts as a machine-learned
second-order Born approximation.
\end{abstract}
\maketitle


\section{Introduction}
Machine Learning (ML) is a highly active field of research that provides a wide range of tools
to tackle various data-based problems. As such, it also receives growing attention in the
theoretical physics literature, such as in Refs.~\cite{Mehta:2014ms,Baldi:2014bsw,Mills:2017,Richards:2011za,Graff:2013cla,Buckley:2011kc,Carleo:2017,Wetzel:2017ooo,He:2017set,Fujimoto:2017cdo,Wu:2018,Niu:2018trk,Brehmer:2018kdj,Steinheimer:2019iso,Larkoski:2017jix}. Many data-based problems involve
modeling an input-target distribution from a data set, which is referred to as supervised learning.
After a successful training procedure, the ML-algorithm is capable of correctly predicting targets,
even when given previously unknown inputs, i.e. it generalizes what it has learned to new data.
Nowadays, neural networks (NNs) are a popular choice in the context of supervised learning. There
is an overwhelming variety of NN-architectures that are as diverse as the problems they are
specially suited for. The certainly most fundamental class of NNs is given by multilayer perceptrons
(MLPs). Many obvious properties of state-of-the-art NNs like the concept of a layered architecture
or the use of non-linear activation functions originate in much simpler MLP-architectures.

The property that makes NNs perform so well in many different applications is that of being an
universal approximator: As long as the architecture comprises an output layer and at least one
hidden layer that is activated via a bounded, non-linear activation function, the NN can approximate
any continuous map between inputs and targets arbitrarily precise for a sufficiently large number
of neurons in that hidden layer, as described by the universal approximation theorem (UAT), see
Refs.~\cite{Cybenko:1989,Hornik:1991}. However, increasing the number of neurons in one layer
is a rather inefficient way to improve the NN's performance. It is more promising to introduce
additional non-linearily activated hidden layers, instead, which eventually opens up the field of
Deep Learning. Here, the term ``deep'' refers to a large number of such non-linearily activated
layers. Its protagonists, the deep neural networks (DNNs), are known for their enormous predictive
power and for demonstrating super-human performances for specific tasks.
Last but not least, this makes them a promising approach towards problems of theoretical physics,
as shown e.g. in Refs.~\cite{Mehta:2014ms,Baldi:2014bsw,Mills:2017}.

\clearpage

While there has clearly been a trend towards deeper and more complex architectures in the past 
decade, capable of approximating increasingly involved target functions, the interest in the 
analytical properties of NNs remains limited to interpretability studies. Notable methods used 
to gather post-hoc interpretations of NN predictions are the deep Taylor decomposition 
and LIME, see Refs.~\cite{Montavon:2017mlbsm} and \cite{Ribeiro:2016rsg}. The former is used 
exclusively for image classifiers: Given a root point of the classifier, a heatmap can be 
constructed using the NN's first-order derivatives, assigning to each pixel a certain relevance 
value. This highlights those pixels, that are substantially involved in the resulting decision.
Similarly, LIME performs a linear regression of adjacent synthetic samples and, thus, provides a 
linear approximation, or in the language of Ref.~\cite{Fan:2020fxw} a linear proxy, of the target 
function in vicinity of the root point. Both methods, indeed, facilitate powerful local 
post-hoc  interpretations of the respective NN's behavior in feature space. However, they do not 
provide any information about second- or higher-order derivatives and are, therefore, blind to
most of the NN's analytical structure. This contrasts greatly with the fact 
a majority of problems in theoretical physics certainly benefit from having not only  
access to the NN's first-order derivatives, but ideally to their entire analytical structure. 
Obvious examples that come to mind are the post-hoc verification of equations of
motion or, what the present study focuses on, the extraction of dominating terms from an 
underlying perturbation theory. 

Closing the mentioned gap requires a general method to compute NN derivatives of arbitrary order. 
On one hand, a numerical differentiation is certainly no difficult task, but may be rendered
inaccurate due to truncation and round-off errors and does not reveal the contribution of the 
individual weights and biases to a particular order. On the other hand, a naive analytical 
differentiation of NN predictions does not share these weaknesses, but will suffer from an 
unmanageable amount of different contributions, especially for high-order derivatives and a large
number of hidden layers. Restricting the analysis for simplicity to MLPs, we propose a 
graph-theoretical formalism to analytically compute partial derivatives of any order for arbitrarily many 
hidden layers, while keeping track of the combinatorics. Similarly to backpropagation in 
gradient-descent techniques, where the first-order derivatives
of the loss function with respect to an internal parameter can be represented as a matrix product,
we want to bypass the naive and inefficient use of the chain- and product-rule and understand
arbitrary derivatives of an MLP in terms of tensor products. We observe two distinct classes of
quantities, we refer to as {\em propagators} and {\em vertices}, that each depend on the weights,
biases and chosen activation functions and naturally appear in such a tensor formulation. Their
naming is intentional, as we discover several similarities between the Taylor expansion of MLPs
and perturbation theory in quantum field theories: Analogously to Feynman rules, we find underlying 
rules that specify which combinations of vertices and propagators, i.e. which diagrams are allowed 
and contribute to a given Taylor coefficient. One major difference, however, is that loops are not 
allowed in contrast to quantum field theories. In a graph-theoretical context, we can show that these
diagrams are oriented and rooted trees, i.e. arborescences. The concept of explaining derivatives 
in terms of graphs is already known and a successful approach in the context of ordinary differential 
equations or, more precisely, the Butcher series, see Ref.~\cite{Hairer:1993hnw}. In contrast to the 
Butcher series, however, we clearly want to set our focus on Taylor expansions and perturbation theory.

Due to its simplicity
and ubiquity in quantum physics, two-body scattering appears to be an adequate field for studying the
adaptation of MLPs to perturbation theories. In fact, the present 
study is strongly motivated by the weight inspections performed in Ref.~\cite{Wu:2018}: Considering the
first hidden layer of MLPs trained to predict S-Wave scattering lengths for shallow, attractive 
potentials of finite range, it is shown that weights among all of its active neurons satisfy a 
quadratic pattern $w_{nm}\propto m^2$. This can be proven to reproduce the first-order Born 
approximation. As soon as MLPs are trained on successively deeper potentials, additional structures 
emerge within their weights, that are later identified with the second-order Born term. This qualitatively
indicates that during training MLPs adapt to the Born series and, thus, develop a quantum perturbation
theory. Applying the proposed graph-theoretical formalism, we complement these findings by a quantitative 
investigation of the dominating analytical structure of MLP ensembles that predict S-wave scattering lengths. 
Since we observe these MLPs to mainly adapt their derivatives to the leading order, we develop an iterative 
scheme that can be understood as an NN perturbation theory to successively obtain remaining terms of the 
target function's Taylor expansion: 
At each iteration, the idea is to eliminate the leading order from the current targets in the training 
and test sets, which generates new data sets for the next iteration, a new auxiliary ensemble of MLPs can 
be trained on. A downside of this approach is that each iteration requires to run an additional training 
pipeline. However, the dominating contributions of the auxiliary ensembles are significantly more faithful 
to the corresponding terms of target functions's Taylor expansion than a differentiation of a single, 
naively trained ensemble could provide. The first- and second-order Taylor coefficients found this way
can be identified one-to-one with the first- and second-order Born term, respectively. These two results are 
then combined to a machine-learned second-order Born approximation, that performs well for shallow
potentials. Finally, these finally indicate that our NN
perturbation theory naturally translates to a perturbation theory for scattering lengths.

The manuscript is organized as follows: At first, Sec.~\ref{sec:varderiv} introduces the 
S-wave scattering length as a functional, briefly presents the first two variational
derivatives and relates them to the Born series in quantum two-body scattering. When approximated 
by NNs, their
sampled form gives rise to understand the Born series as a Taylor series in the space of sampled 
potentials. 
In Sec.~\ref{sec:pt}, we then propose a graphically motivated approach to compute partial derivatives 
of arbitrarily deep MLPs
in terms of propagators and vertices. 
Sec.~\ref{sec:network} builds on these findings, develops the NN perturbation theory we finally want 
to apply to MLPs trained on S-wave scattering lengths and sheds light on the training pipeline as 
well as on architecture details. 
The first-order Born term is evaluated and
discussed in Sec.~\ref{sec:born1}, followed by the investigation of the second-order Born term after 
one iteration
in Sec.~\ref{sec:born2}. We end with a discussion and outlook in Sec.~\ref{sec:summ}. Various
technicalities are relegated to the appendix.
\vfill\noindent

\section{The Born series as a Taylor series}
\label{sec:varderiv}
Let $\bm{Y}:\mathbb{R}^{H_0}\to\mathbb{R}^{H_L}$ be an analytical function. The local behavior of $\bm{Y}$ 
in the vicinity of an arbitrary point $\bm{x}_0\in\mathbb{R}^{H_0}$ can be described by neglecting 
higher-order terms in the Taylor expansion
\vfill\noindent
\begin{equation}
\begin{aligned}
Y_n(\bm{x}) - Y_n(\bm{x}_0) &= \sum\limits_{k=1}^{H_0} \left. \frac{\partial Y_n}{\partial x_{k}} \right|_{\bm{x}=\bm{x}_0} (\bm{x}-\bm{x}_0)_{k} \\&+ \frac{1}{2}\sum\limits_{k_1=1}^{H_0}\sum\limits_{k_2=1}^{H_0} \left. \frac{\partial^2 Y_n}{\partial x_{k_1} \partial x_{k_2}} \right|_{\bm{x}=\bm{x}_0}(\bm{x}-\bm{x}_0)_{k_1}(\bm{x}-\bm{x}_0)_{k_2} \\&+ \ldots
\end{aligned} \label{eq-taylor-x3}
\end{equation}
\vfill\noindent
In the following, we interpret the components $x_k=f(k/H_0)$ of any given vector 
$\bm{x}\in\mathbb{R}^{H_0}$ as samples of an analytical function ${f\in C^\omega([0,1])}$. 
In this context, higher dimensions~$H_0$ correspond to higher sampling 
rates. Note that $\bm{x}$ and $f$ become totally equivalent in the limit of an infinitely fine 
sampling rate, that is $H_0\to\infty$. In this case the given function $\bm{Y}$ can rather be 
understood as a functional $\bm{Y}:C^\omega([0,1])\to \mathbb{R}^{H_L}$. Consequently, the above 
expression generalizes to an expansion around a given function $f_0\in C^\omega([0,1])$,
\vfill\noindent
\begin{align}
Y_n[f] - Y_n[f_0] = 
&\ \int_0^1 \! \mathrm{d}k \, \left. \frac{\delta Y_n[f]}{\delta f(k)} \right|_{f=f_0} \{f(k)-f_0(k)\} \notag\\
&+ \frac{1}{2}\int_0^1 \! \mathrm{d}k_1 \, \int_0^1 \! \mathrm{d}k_2 \, \left. \frac{\delta^2 Y_n[f]}{\delta f(k_1) \delta f(k_2)} \right|_{f=f_0} \{f(k_1)-f_0(k_1)\}\{f(k_2)-f_0(k_2)\} \notag\\
&+ \ldots \notag
\end{align} 
\clearpage\noindent
In this limit it is no longer the partial derivatives 
$\partial^N \bm{Y}/\partial x_{k_1}\ldots\partial x_{k_N}|_{\bm{x}=\bm{x}_0}$, but the variational 
derivatives $\delta^N \bm{Y}[f]/\delta f(k_1)\ldots\delta f(k_N)|_{f=f_0}$ that parametrize the 
local behavior of $\bm{Y}$. Sampling the latter yields partial derivatives and again reproduces the 
multidimensional Taylor expansion in Eq.~\eqref{eq-taylor-x3}. An example we study thoroughly in 
Sec.~\ref{sec:born1} and Sec.~\ref{sec:born2} is the functional that maps an attractive, 
dimensionless potential ${U=2\mu\rho^2 V}$ with finite range $\rho$ to the corresponding 
dimensionless S-wave scattering length in units of $\rho$,
\vfill\noindent
\begin{equation}
\begin{aligned}
a_0[U] &= \frac{2\pi^2}{\rho^3} \braket{0|T|0} = \frac{2\pi^2}{\rho^3} \braket{0|U|0} + \frac{\pi^2}{\rho^6} \braket{0|U G_0 U|0} + \ldots \\
&= \int_0^1 \! \mathrm{d}r \, r^2 U(r) - \frac{1}{2}\int_0^1 \! \mathrm{d}r_1 \,
\int_0^1 \! \mathrm{d}r_2 \, r_1r_2(r_1+r_2-|r_1-r_2|) U(r_1)U(r_2)
+ \ldots
\end{aligned} \label{eq-born-1}
\end{equation}
\vfill\noindent
Here, the quantities $\mu$, $V$, $T$ and $G_0$ denote the reduced mass of the two-body system, the 
dimensionful potential and the dimensionless T-matrix and resolvent, respectively. 
Eq.~\eqref{eq-born-1} not only contains the expansion of $a_0$ around the force-free case $U=0$: It 
also displays the classical representation of the S-wave scattering length as the Born series and, 
therefore, suggests to treat $a_0$ perturbatively for shallow potentials. The two lowest-order 
variational derivatives,
\vfill\noindent
\begin{equation}
\left. \frac{\delta a_0[U]}{\delta U(r)} \right|_{U=0} = r^2, \hspace{1cm} \left. \frac{\delta^2 a_0[U]}{\delta U(r_1) \delta U(r_2)} \right|_{U=0} = - r_1r_2(r_1+r_2-|r_1-r_2|), \notag
\end{equation}
\vfill\noindent
can then be used to compute the first and, respectively, second-order Born approximation of $a_0$. 
Consequently, their sampled versions are given by
\vfill\noindent
\begin{equation}
\left.\frac{\partial a_0}{\partial U_k} \right|_{\bm{U}=\bm{0}} = \frac{k^2}{(H_0)^3}, \hspace{1cm} \left.\frac{\partial^2 a_0}{\partial U_{k_1} \partial U_{k_2}} \right|_{\bm{U}=\bm{0}} = -\frac{1}{(H_0)^5}k_1k_2(k_1+k_2-|k_1-k_2|). \label{eq-born-kernel-x3}
\end{equation}
\vfill\noindent
Eqs.~\eqref{eq-taylor-x3} and \eqref{eq-born-kernel-x3} give rise to understand the Born series in 
the space of sampled potentials as a Taylor series. Thereby, each sampled potential $\bm{U}$ 
corresponds to a $H_0$-dimensional vector with components ${U_k = U(r=k/H_0)}$. It is obvious that 
the discretization error becomes negligibly small for sufficiently high sampling rates $H_0$.
Now $\bm{Y}$ could serve as a target function that we try to imitate by an NN. In the context of the 
above example this means that the NN can successfully predict S-wave scattering lengths for sampled 
potentials after completing a supervised training procedure. According to the UAT, these predictions 
can be arbitrarily precise, provided that the given architecture contains sufficiently many neurons 
or is sufficiently deep. Nonetheless, the UAT in no way guarantees that the NN also reproduces the 
analytical properties, i.e. the target function's partial derivatives at each order. A pathological, but obvious 
example are NNs with Heaviside-activations: Here the derivatives at any order can only take the values $0$ or 
$\pm\infty$, which can be realized as a superposition of delta functions.

At this point the following questions arise: What conditions must the given architecture satisfy such 
that loss minimization during training also causes the partial derivatives of the MLP for any given order to approximate 
the corresponding derivatives of the target function? Or asked differently: If we are given a raw data 
set and do not know the analytical representation of the target function, how can we be sure that its 
analytical structure is reproduced by the trained NN? What is the benefit of having the NN approximate 
the analytical structure of the target function? And how can MLP and target function derivatives be 
compared with each other analytically in the first place?

\clearpage

Of course, in order to comply with the assumed analyticity of the target function, the activation 
functions themselves need to be analytical. As many pooling layers like Max-Pooling or rectifiers like ReLU 
are not everywhere differentiable, this already excludes a wide range of conventional architectures. 
Also, note that this analyticity criterion is just a neccessary and not a sufficient condition for the 
NN in order to approximate the analytical structure of the target function. To give an example, GELU is 
an analytical rectifier that serves as activation function later in this work. While lower-order derivatives 
are unproblematic, higher-order derivatives vanish almost everywhere and diverge for a small range of 
inputs, which renders them highly unstable. Also, there is no ad-hoc guarantee, that the NN approximates 
all lower-order derivatives of the target function equally well: If the training set only covers a 
narrow range of inputs around the expansion point, the NN may tend to approximate only the leading order. 
While we will observe this issue for the numerically stable first- and second-order derivatives of the Born 
series in Sec.~\ref{sec:born1} and \ref{sec:born2}, a further 
investigation for higher-order derivatives is beyond the scope of the present study and, therefore, 
left for future studies.

\vfill\noindent

\section{Partial Derivatives of Multilayer Perceptrons}
\label{sec:pt}

An MLP with $L$ layers is the prototype example of a layered architecture and can be understood
as a non-linear function $\bm{\mathcal{Y}}:\mathbb{R}^{H_0}\to\mathbb{R}^{H_L}$ between real vector spaces. 
The
term ``layered'' describes that $\bm{\mathcal{Y}}$ is a composition 
$\bm{\mathcal{Y}}=\bm{\mathcal{Y}}_L\circ\ldots \circ\bm{\mathcal{Y}}_1$ of $L$ layers
$\bm{\mathcal{Y}}_l:\mathbb{R}^{H_{l-1}}\to\mathbb{R}^{H_l}$ containing $H_l$ neurons $z_n^{(l)}$. 
In MLPs exclusively
linear layers are used in combination with non-linear activation functions
$a^{(l,n)}:\mathbb{R}\to\mathbb{R}$, where $a^{(l,n)}$ is applied to the $n^\text{th}$ neuron of the
$l^\text{th}$ layer. This can be formulated recursively,
\vfill\noindent
\begin{equation}
\mathcal{Y}_n(\bm{x})=y_n^{(L)},
\end{equation}
\vfill\noindent
with the recursive step
\vfill\noindent
\begin{equation}
y_n^{(l)}=a^{(l,n)}\left(z_n^{(l)}\right), \hspace{0.6cm} z_n^{(l)} = \sum\limits_{m=1}^{H_{l-1}}
w_{nm}^{(l)} y_m^{(l-1)}+\ b_n^{(l)}, \label{eq-mlprecursion}
\end{equation}
\vfill\noindent
together with the weights $w_{nm}^{(l)}$ and biases $b_n^{(l)}$. The recursion in
Eq.~\eqref{eq-mlprecursion} is terminated for $l=1$ due to reaching its base $y_m^{(0)}=x_m$.
For deep architectures, that is for $L\gg 1$, $\bm{\mathcal{Y}}$ is a strongly nested function, such 
that computing
derivatives becomes an extremely difficult task because of a hardly managable amount of chain-
and product-rule applications. In fact, there is another field of machine learning in which it
is well known how to efficiently compute first order partial derivatives of a strongly nested
function: Within gradient-descent-based training algorithms it is necessary to compute the gradient
of a loss function, that is an error function of the network~$\bm{\mathcal{Y}}$ and therefore nested to the
same extent. Here, the first order partial derivatives of the loss function with respect to any
internal parameter can be expressed by a matrix product. This is the famous backpropagation which
significantly speeds up training steps by avoiding to naively apply chain- and product-rules, see Ref.~\cite{Nielsen:2015}.

\vfill

In order to derive Taylor coefficients of $\bm{\mathcal{Y}}$ of any order and for an arbitrary number $L$ of
layers in terms of the weights and biases, we desire a systematic description in the spirit of
backpropagation. Let us therefore at first define 

\vfill\noindent
\begin{equation}
D_{nm}^{(l,p)}=w_{nm}^{(l+1)}\frac{{\mathrm{d}\vphantom{d^{d}}}^p a^{(l,m)}}{{\mathrm{d}x}^p}
\left(z_m^{(l)}\right),\label{eq-prop-1}
\end{equation}   
\clearpage
\noindent
which we refer to as the $nm^\text{th}$ matrix element of the $l^\text{th}$ layer {\em propagator}
of order $p$. Since the last layer is usually activated via the identity, $a^{(L,n)}=\mathrm{id}$,
and has no bias, that is $b_{n}^{(L)}=0$, we can write
\vfill
\noindent
\begin{equation}
\mathcal{Y}_n(\bm{x})=\sum\limits_{m=1}^{H_{L-1}}D_{nm}^{(L-1,0)}.
\end{equation}
\vfill
\noindent
This redefinition entirely describes outputs in terms of propagators and reduces the search
for Taylor coefficients to computing partial derivatives of propagator matrix elements,
\vfill
\noindent
\begin{equation}
\frac{\partial^N \mathcal{Y}_n}{\partial x_{k_1}\ldots \partial x_{k_N}} = \sum\limits_{m=1}^{H_{L-1}} \frac{\
 \hspace{0.5cm} \partial^N D_{nm}^{(L-1,0)}}{\partial x_{k_1}\ldots \partial x_{k_N}}. \label{eq-yderiv-1}
\end{equation}
\vfill
\noindent
\begin{equation}
\frac{\partial D_{nm}^{(l,p)}}{\partial x_k} = D_{nm}^{(l,p+1)} \Delta_{mk}^{(l,1)}, \hspace{1cm}
\Delta_{mk}^{(l,1)}{=\frac{\partial z_{m}^{(l)}}{\partial x_k}}=\sum\limits_{q_l=1}^{H_l}\ldots\sum\limits_{q_1=1}^{H_1}\delta_{mq_l} w_{q_1 k}^{(1)}
\prod\limits_{i=1}^{l-1}D_{q_{i+1}q_{i}}^{(i,1)}. \label{eq-propderiv-1}
\end{equation}
\vfill
\noindent
In Eq.~\eqref{eq-propderiv-1} we make two observations: First, a derivation increases the order
of the propagator by one. Second, an additional factor $\Delta_{mk}^{(l,1)}$ is introduced, that
impacts higher order derivatives of propagators. We introduce the tensor elements
\vfill
\noindent
\begin{equation}
{\Delta_{mk_1\ldots k_{p}}^{(l,p)}=\frac{\partial z_{m}^{(l)}}{\partial x_{k_{1}} \ldots \partial x_{k_{p}}},}
\label{eq-delta-1}
\end{equation}
\vfill
\noindent
that are obviously invariant under permutations of the indices $k_1,\ldots,k_p$. Now we can express the 
$N^\text{th}$ derivative of the propagator as the following superposition
by successively applying the rule mentioned in Eq.~\eqref{eq-propderiv-1} and by absorbing the
remaining derivatives by Eq.~\eqref{eq-delta-1} (see App.~A on p.~\pageref{sec-proofs}),
\vfill
\noindent
\begin{equation}
\frac{\ \hspace{0.1cm}\partial^N D_{nm}^{(l,p)}}{\partial x_{k_1}\ldots\partial x_{k_N}} = \sum\limits_{c=1}^ND_{nm}^{(l,p+c)}\hspace{-0.2cm}\sum\limits_{\sigma\in\mathcal{S}_N}\sum\limits_{\bm{\pi}\in \Pi^c_N}\frac{1}{\varepsilon_{\bm{\pi}}}\prod\limits_{i=1}^c \Delta_{mk_{\sigma\left(1+\sum_{j=1}^{i-1}\pi_j\right)}\ldots\ k_{\sigma\left(\sum_{j=1}^{i}\pi_j\right)}}^{(l,\pi_i)}. \label{eq-propderiv-2}
\end{equation}
\vfill
\noindent
Each summand in Eq.~\eqref{eq-propderiv-2} includes a higher order propagator. Here, the
second sum runs over the set of all partitions of the number $N$ with length $c$,
\vfill
\noindent
\begin{equation}
\Pi^c_N = \left\{\bm{\pi}\in\mathbb{N}^c \left| \sum\limits_{i=1}^c \pi_i = N \right.
\wedge \left(\pi_1 \geq \ldots \geq \pi_c\right) \right\}, \hspace{1cm} \Pi_N=\bigcup\limits_{c=1}^N \Pi_N^c. \label{eq-part-def}
\end{equation}
\clearpage
\begin{table}[t]
\begin{minipage}{\textwidth} 
\resizebox{\textwidth}{!}{
\renewcommand{\arraystretch}{2}
\begin{tabular}{||clll||} \hhline{|t:====:t|}
\hspace{0.5cm}$N$\hspace{0.5cm} & $\bm{\pi}\in\Pi_N$\hspace{0.5cm} & $\varepsilon_{\bm{\pi}}$\hspace{0.5cm} & \scalebox{1.3}{$\frac{\ \hspace{0.1cm}\partial^N D_{nm}^{(l,p)}}{\partial x_{k_1}\ \ldots\ \partial x_{k_N}}$}\hspace{0.5cm} \\ \hhline{|:====:|}
$1$ & $(1)$ & $1$ & $D_{nm}^{(l,p+1)}\Delta_{mk_1}^{(l,1)}$ \\ \hhline{|:====:|}
\multirow{2}{*}{$2$} & $(1,1)$ & $2$ & $D_{nm}^{(l,p+2)} \Delta_{mk_1}^{(l,1)}\Delta_{mk_2}^{(l,1)}$ \\
& $(2)$ & $2$ & $+ D_{nm}^{(l,p+1)}\Delta_{mk_1k_2}^{(l,2)}$ \\ \hhline{|:====:|}
\multirow{3}{*}{$3$} & $(1,1,1)$ & $6$ & $D_{nm}^{(l,p+3)}\Delta_{mk_1}^{(l,1)}\Delta_{mk_2}^{(l,1)}\Delta_{mk_3}^{(l,1)}$ \\
& $(2,1)$ & $2$ & $+ D_{nm}^{(l,p+2)}\left(\Delta_{mk_1k_2}^{(l,2)}\Delta_{mk_3}^{(l,1)}+\Delta_{mk_1k_3}^{(l,2)}\Delta_{mk_2}^{(l,1)}+\Delta_{mk_2k_3}^{(l,2)}\Delta_{mk_1}^{(l,1)} \right)$ \\
& $(3)$ & $6$ & $+ D_{nm}^{(l,p+1)} \Delta_{mk_1k_2k_3}^{(l,3)}$ \\ \hhline{|:====:|}
\multirow{6}{*}{$4$} & $(1,1,1,1)$ & $24$ & $D_{nm}^{(l,p+4)}\Delta_{mk_1}^{(l,1)}\Delta_{mk_2}^{(l,1)}\Delta_{mk_3}^{(l,1)}\Delta_{mk_4}^{(l,1)}$ \\
& \multirow{2}{*}{$(2,1,1)$} & \multirow{2}{*}{$4$} &  $+D_{nm}^{(l,p+3)}\left( \Delta_{mk_1k_2}^{(l,2)}\Delta_{mk_3}^{(l,1)}\Delta_{mk_4}^{(l,1)}+\Delta_{mk_1k_3}^{(l,2)}\Delta_{mk_2}^{(l,1)}\Delta_{mk_4}^{(l,1)} +\Delta_{mk_1k_4}^{(l,2)}\Delta_{mk_2}^{(l,1)}\Delta_{mk_3}^{(l,1)}\right. $ \\
&&& $\hspace{1.55cm}\left. +\Delta_{mk_2k_3}^{(l,2)}\Delta_{mk_1}^{(l,1)}\Delta_{mk_4}^{(l,1)} + \Delta_{mk_2k_4}^{(l,2)}\Delta_{mk_1}^{(l,1)}\Delta_{mk_3}^{(l,1)} +\Delta_{mk_3k_4}^{(l,2)}\Delta_{mk_1}^{(l,1)}\Delta_{mk_2}^{(l,1)} \right)$ \\
& $(2,2)$ & $8$ & $+D_{nm}^{(l,p+2)} \left( \Delta_{mk_1k_2}^{(l,2)}\Delta_{mk_3k_4}^{(l,2)}+\Delta_{mk_1k_3}^{(l,2)}\Delta_{mk_2k_4}^{(l,2)}+\Delta_{mk_1k_4}^{(l,2)}\Delta_{mk_2k_3}^{(l,2)} \right)$ \\
& $(3,1)$ & $6$ & $+D_{nm}^{(l,p+2)} \left( \Delta_{mk_1k_2k_3}^{(l,3)}\Delta_{mk_4}^{(l,1)}+\Delta_{mk_1k_2k_4}^{(l,3)}\Delta_{mk_3}^{(l,1)}+\Delta_{mk_1k_3k_4}^{(l,3)}\Delta_{mk_2}^{(l,1)}+\Delta_{mk_2k_3k_4}^{(l,3)}\Delta_{mk_1}^{(l,1)} \right)$ \\
& $(4)$ & $24$ & $+D_{nm}^{(l,p+1)}\Delta_{mk_1k_2k_3k_4}^{(l,4)}$ \\ \hhline{|b:====:b|}
\end{tabular}
}
\caption{All partitions $\bm{\pi}\in\Pi_N$ and corresponding factors $\varepsilon_{\bm{\pi}}$ for $N=1, 2, 3, 4$ required for computing the propagator
derivatives $\partial^N D_{nm}^{(l,p)}/\left(\partial x_{k_1}\ \ldots\ \partial x_{k_N}\right)$ using
Eq.~\eqref{eq-propderiv-2}. Note that the index permutation symmetry of the $\Delta_{mk_1\ldots k_{p}}^{(l,p)}$ has been used to combine $\varepsilon_{\bm{\pi}}$ summands. Thus, there are only $N!/\varepsilon_{\bm{\pi}}$ remaining summands for each partition $\bm{\pi}$.}\label{tab-partperm}
\end{minipage}
\end{table}
The set $\Pi_N$ of all partitions is, thereby, simply given by the union of all $\Pi_{N}^{c}$.
Lastly, the third sum runs over the permutation group $\mathcal{S}_N$. For a given partition $\bm{\pi}\in\Pi_{n}^{c}$, the factor $1/\varepsilon_{\bm{\pi}}$ takes the symmetry of the tensor elements $\Delta_{mk_{\sigma(1)}\ldots k_{\sigma(p)}}^{(l,p)}=\Delta_{mk_1\ldots k_{p}}^{(l,p)}$ under index permutations $\sigma\in\mathcal{S}_N$ into account and can be derived via
\vspace{-0.5cm}
\begin{equation}
\varepsilon_{\bm{\pi}}=\prod\limits_{i=1}^{c}(\pi_i)!\left[\left(\sum\limits_{j=1}^{c}\delta_{\pi_i\pi_j}\right)!\right]^{1/\sum\limits_{j=1}^{c}\delta_{\pi_i\pi_j}}.
\end{equation}
\noindent
Tab.~\ref{tab-partperm} contains all partitions, respective $\varepsilon_{\bm{\pi}}$
and resulting propagator derivatives for $N=1,2,3,4$. What remains is to find an expression of the tensors $\Delta_{mk_1\ldots k_{p}}^{(l,p)}$ in terms of
propagators such that we can completely determine the partial derivatives of the propagators
in Eq.~\eqref{eq-propderiv-2}. We therefore introduce the $mk^\text{th}$ matrix element of a
vertex of order $p$ in the $l^\text{th}$ layer, acting as a weighted sum over elements of arbitrary tensors $f$,
\vfill
\noindent
\begin{equation}
\mlpsum\limits_{(q_a)_{a=1}^{l} \vphantom{\sum\limits^,} (j_b)_{b=1}^{p}}\hspace{-0.7cm}{\vphantom{\mlpsum}}_{mk}
\hspace{0.2cm} f_{q_{j_1}\ldots q_{j_p}} = \sum\limits_{q_l}^{H_l}\ldots\sum\limits_{q_1}^{H_1}\delta_{mq_l}
w_{q_1 k}^{(1)}\sum\limits_{j_1=1}^{l-1}\sum\limits_{\mbox{\scriptsize $ \begin{matrix}j_2=1\\ j_2\neq j_1
\end{matrix} $ }}^{l-1}\ldots\sum\limits_{\mbox{\scriptsize $ \begin{matrix}j_p=1\\ j_p \neq j_1 \\
      \vdots \\ j_p \neq j_{p-1} \end{matrix} $ }}^{l-1}\left( \vphantom{\prod\limits_i^i} \right.
\prod\limits_{\mbox{\scriptsize $ \begin{matrix}i=1\\ i\neq j_1 \\ \vdots \\ i\neq j_p \end{matrix} $ }}^{l-1}
D_{q_{i+1}q_{i}}^{(i,1)} \left. \vphantom{\prod\limits_i^i} \right) f_{q_{j_1}\ldots q_{j_p}}.
\end{equation}
If $l-1\leq p$, the vertex becomes saturated, that is it becomes a constant and is equal to any
higher order vertex in the same layer. Because of this and due to Eq.~\eqref{eq-propderiv-1},
vertices display the following behavior when exposed to a partial derivative,
\vfill
\noindent
\begin{align}
\frac{\partial}{\partial x_{k_2}}\mlpsum\limits_{(q_a)_{a=1}^{l} \vphantom{\sum\limits^,} (j_b)_{b=1}^{p-1}}\hspace{-0.7cm}{\vphantom{\mlpsum}}_{mk_1}\hspace{0.2cm} f_{q_{j_1}\ldots q_{j_{p-1}}} = \Theta(l-p) \mlpsum\limits_{(q_a)_{a=1}^{l} \vphantom{\sum\limits^,} (j_b)_{b=1}^{p}}\hspace{-0.7cm}{\vphantom{\mlpsum}}_{mk_1}\hspace{0.2cm} f_{q_{j_1}\ldots q_{j_{p-1}}} D_{q_{j_{p+1}}q_{j_{p}}}^{(j_p,2)} \mlpsum\limits_{(q_a^\prime)_{a=1}^{j_p} \vphantom{\sum\limits^,}}\hspace{-0.2cm}{\vphantom{\mlpsum}}_{q_{j_{p}}k_2} \notag \\ +\mlpsum\limits_{(q_a)_{a=1}^{l} \vphantom{\sum\limits^,} (j_b)_{b=1}^{p-1}}\hspace{-0.7cm}{\vphantom{\mlpsum}}_{mk_1}\hspace{0.2cm} \frac{\partial}{\partial x_{k_2}} f_{q_{j_1}\ldots q_{j_{p-1}}}.\label{eq-vertexderiv-1}
\end{align}
\vfill
\noindent
Obviously, vertices of order $p$ in the $l^\text{th}$ layer only commute with partial derivatives, if
they are saturated. This is embodied by the proportionality of the commutator to the step-function with 
$\Theta(0)=0$. Note that we have expressed
$\Delta_{mk}^{(l,1)}$ as a vertex of order zero in order to arrive at Eq.~\eqref{eq-vertexderiv-1},
\vfill
\noindent
\begin{align}
\Delta_{mk_1}^{(l,1)} = \mlpsum\limits_{(q_a)_{a=1}^{l} \vphantom{\sum\limits^,}}\hspace{-0.2cm}
{\vphantom{\mlpsum}}_{mk_1} = 
\Theta(l-0)\left(\ \begin{tikzpicture}[baseline=-2pt, inner sep=0pt, outer sep=4pt]
    \begin{feynman}
        \vertex [dot] (k1) at (0,0) {};
        \diagram*{
            (k1) -- (k1)
        };
    \end{feynman}
\end{tikzpicture}\vphantom{\sum}\ \right)_{mk_1}^{(l,1)}, \label{eq-vertex-1}
\end{align}
\vfill
\noindent
for which we choose the graphical representation of a single vertex. Applying
Eq.~\eqref{eq-vertexderiv-1} to Eq.~\eqref{eq-vertex-1} yields
\vfill
\noindent
\begin{align}
\Delta_{mk_1k_2}^{(l,2)}&=\Theta(l-1)\mlpsum\limits_{(q_a)_{a=1}^{l} \vphantom{\sum\limits^,} (j_1)}\hspace{-0.45cm}{\vphantom{\mlpsum}}_{mk_1} D_{q_{j_1+1}q_{j_1}}^{(j_1,2)}\mlpsum\limits_{(q_a^\prime)_{a=1}^{j_1} \vphantom{\sum\limits^,}}\hspace{-0.2cm}{\vphantom{\mlpsum}}_{q_{j_1}k_2} \notag \\&= \Theta(l-1)\Theta(l-0)\mlpsum\limits_{(q_a)_{a=1}^{l} \vphantom{\sum\limits^,} (j_1)}\hspace{-0.45cm}{\vphantom{\mlpsum}}_{mk_1} D_{q_{j_1+1}q_{j_1}}^{(j_1,2)} \left(\ \begin{tikzpicture}[baseline=-2pt, inner sep=0pt, outer sep=4pt]
    \begin{feynman}
        \vertex [dot] (k1) at (0,0) {};
        \diagram*{
            (k1) -- (k1)
        }; 
    \end{feynman}
\end{tikzpicture}\vphantom{\sum}\ \right)_{q_{j_1}k_2}^{(j_1,1)} \notag \\
&= \Theta(l-1)\left( \begin{tikzpicture}[baseline=-2pt, inner sep=0pt, outer sep=4pt]
    \begin{feynman}
        \vertex [dot] (k1) at (0,0) {};
        \vertex [dot] (k2) at (1,0) {};
        \diagram*{
            (k1) -- [dn] (k2)
        };
    \end{feynman}
\end{tikzpicture}\vphantom{\sum}\right)_{mk_1k_2}^{(l,2)}~.\label{eq-deltak1k2-x1}
\end{align}
\vfill
\noindent
This term depends on a first order vertex that sums over a second order propagator and a zeroth
order vertex, which suggests the graphical representation of two vertices that are connected
via a propagator with one arrow head, directed from the first to the second vertex. Note that the 
index permutation symmetry of the $\Delta^{(l,p)}_{mk_1\ldots k_p}$ is inherited by the right-hand 
side of Eq.~\eqref{eq-deltak1k2-x1}. Applying Eq.~\eqref{eq-vertexderiv-1} once more to 
Eq.~\eqref{eq-deltak1k2-x1} yields
\vfill
\noindent
\begin{align}
\Delta_{mk_1k_2k_3}^{(l,3)}=
\ &\Theta(l-2)\hspace{-0.15cm}\mlpsum\limits_{(q_a)_{a=1}^{l} \vphantom{\sum\limits^,} (j_1,j_2)}\hspace{-0.6cm}{\vphantom{\mlpsum}}_{mk_1}D_{q_{j_1+1}q_{j_1}}^{(j_1,2)}D_{q_{j_2+1}q_{j_2}}^{(j_2,2)}\mlpsum\limits_{(q_a^\prime)_{a=1}^{j_1} \vphantom{\sum\limits^,}}\hspace{-0.2cm}{\vphantom{\mlpsum}}_{q_{j_1}k_2}\mlpsum\limits_{(q_a^{\prime\prime})_{a=1}^{j_2} \vphantom{\sum\limits^,}}\hspace{-0.2cm}{\vphantom{\mlpsum}}_{q_{j_2}k_3}\notag\\
+\, &\Theta(l-1) \mlpsum\limits_{(q_a)_{a=1}^{l} \vphantom{\sum\limits^,} (j_1)}\hspace{-0.45cm}{\vphantom{\mlpsum}}_{mk_1}D_{q_{j_1+1}q_{j_1}}^{(j_1,3)}\mlpsum\limits_{(q_a^\prime)_{a=1}^{j_1} \vphantom{\sum\limits^,}}\hspace{-0.2cm}{\vphantom{\mlpsum}}_{q_{j_1}k_2}\mlpsum\limits_{(q_a^{\prime\prime})_{a=1}^{j_1} \vphantom{\sum\limits^,}}\hspace{-0.2cm}{\vphantom{\mlpsum}}_{q_{j_1}k_3}\notag\\
+\, &\Theta(l-1)\mlpsum\limits_{(q_a)_{a=1}^{l} \vphantom{\sum\limits^,} (j_1)}\hspace{-0.45cm}{\vphantom{\mlpsum}}_{mk_1}D_{q_{j_1+1}q_{j_1}}^{(j_1,2)}\mlpsum\limits_{(q_a^\prime)_{a=1}^{j_1} \vphantom{\sum\limits^,}(j_1^\prime)}\hspace{-0.45cm}{\vphantom{\mlpsum}}_{q_{j_1}k_2}D_{q^\prime_{j_1^\prime+1}q_{j_1^\prime}}^{(j_1^\prime,2)}\mlpsum\limits_{(q_a^{\prime\prime})_{a=1}^{j_1} \vphantom{\sum\limits^,}}\hspace{-0.2cm}{\vphantom{\mlpsum}}_{q^\prime_{j_1^\prime}k_3},\notag
\end{align}
\vfill
\noindent
which can be graphically represented as
\vfill
\noindent
\begin{align}
\Delta_{mk_1k_2k_3}^{(l,3)} =\ & \Theta(l-2)\left( \begin{tikzpicture}[baseline=-2pt, inner sep=0pt, outer sep=4pt]
    \begin{feynman}
        \vertex [dot] (k1) at (0,0) {};
        \vertex [dot] (k2) at (0.87,0.5) {};
        \vertex [dot] (k3) at (0.87,-0.5) {};
        \diagram*{
            (k1) -- [dn] (k2),
            (k1) -- [d2] (k3)
        };
    \end{feynman}
\end{tikzpicture}\ \right)_{mk_1k_2k_3}^{(l,3)}
+
\Theta(l-1)\left( \begin{tikzpicture}[baseline=-2pt, inner sep=0pt, outer sep=4pt]
    \begin{feynman}
        \vertex [dot] (k1) at (0,0) {};
        \vertex [dot] (k2) at (0.87,0.5) {};
        \vertex [dot] (k3) at (0.87,-0.5) {};
        \diagram*{
            (k1) -- [dn] (k2),
            (k1) -- [dn] (k3)
        };
    \end{feynman}
\end{tikzpicture}\ \right)_{mk_1k_2k_3}^{(l,3)}
+
\Theta(l-1)\left( \begin{tikzpicture}[baseline=-2pt, inner sep=0pt, outer sep=4pt]
    \begin{feynman}
        \vertex [dot] (k1) at (0,0) {};
        \vertex [dot] (k2) at (0.87,0.5) {};
        \vertex [dot] (k3) at (0.87,-0.5) {};
        \diagram*{
            (k3) -- [dn] (k1) -- [dn] (k2)
        };
    \end{feynman}
\end{tikzpicture}\ \right)_{mk_1k_2k_3}^{(l,3)} . \label{eq-deltak1k2k3-x1}
\end{align}
\clearpage
This is a superposition of three different terms, to each of them we can  assign a different graph. Note 
that the second term only contains one propagator of third order instead of two second-order 
propagators as in both other terms. Given a vertex, we decide to enumerate outgoing propagators by 
the number of their arrow heads. If there are $n$ outgoing edges with the same number of arrow heads, 
as it is the case in the second term with $n=2$, they represent the same propagator of order $n+1$.

It would not be of much use to continue with successively deriving
higher order tensors as above. The general idea of how the $\Delta_{mk_1\ldots k_{N}}^{(l,N)}$ are
structured and how the individual terms can be translated into graphs should be clear: 
\begin{itemize}
\item $\Delta_{mk_1\ldots k_{N}}^{(l,N)}$ can be represented by a sum of directed and rooted trees, 
that is arborescences, see Ref.~\cite{Kamiyama:2014k}. Each arborescence 
consists of $N$ vertices and up to $N-1$ propagators.
\item At each vertex, outgoing propagators are counted by the number of arrow heads on
the respective edges. Therefore, an edge with $n$ arrow heads belongs to the $n^\text{th}$ propagator 
originating at that vertex.
\item If there are $n$ edges with the same amount of arrow heads originating in a given vertex, these 
represent a propagator of
order $n+1$ originating in that vertex. Consequently, this propagator establishes connections to $n$ 
other vertices.
	\item A term of the structure
	\begin{equation*} 
	\ldots\mlpsum\limits_{(q_a^{\{1\}})_{a=1}^{j_1^{\{0\}}} \vphantom{\begin{matrix}{}\\{}\\{}\end{matrix}} (j_a^{\{1\}})_{a=1}^{p_1}}\hspace{-0.9cm}{\vphantom{\mlpsum}}_{q^{\{0\}}_{j_1^{\{0\}}}k_{x_1}} D_{q^{\{1\}}_{j^{\{1\}}_b+1}q_{j^{\{1\}}_b}}^{(j^{\{1\}}_b,n)}\ldots\mlpsum\limits_{(q_a^{\{2\}})_{a=1}^{j_a^{\{1\}}} \vphantom{\begin{matrix}{}\\{}\\{}\end{matrix}} (j_a^{\{2\}})_{a=1}^{p_2}}\hspace{-0.9cm}{\vphantom{\mlpsum}}_{q^{\{1\}}_{j^{\{b\}}_1}k_{x_2}}\ldots\mlpsum\limits_{(q_a^{\{n\}})_{a=1}^{j_a^{\{1\}}} \vphantom{\begin{matrix}{}\\{}\\{}\end{matrix}} (j_a^{\{n\}})_{a=1}^{p_n}}\hspace{-0.9cm}{\vphantom{\mlpsum}}_{q^{\{1\}}_{j^{\{b\}}_1}k_{x_n}}\ldots\vspace{-0.7cm}
	\end{equation*}
	indicates that it is the $b^\text{th}$ propagator originating in the $k_{x_1}{}^\text{st}$
        vertex that establishes a connection to the $k_{x_2}{}^\text{nd},\ldots,k_{x_{n-1}}{}^\text{th}$
        and $k_{x_n}{}^\text{th}$ vertex. This implies that this propagator is of order $n$. If no
        propagator is originating in a vertex, that vertex is called a leaf of the given arborescence.
	\item Derivatives of saturated vertices vanish, as seen in Eq.~\eqref{eq-vertexderiv-1}. Thus, it 
	depends on the layer $l$, propagators and vertices are considered for, whether a certain arborescence 
	contributes or not. This is embodied by multiplying a factor $\Theta(l-\alpha)$ to each arborescence. 
	The given arborescence only contributes, if $l$ overshoots its saturation threshold, which we denote 
	by $\alpha$. The appearance of internal vertices and propagators of order three or higher decreases 
	$\alpha$.
\end{itemize}
In order to pursue a more systematical approach, we want to understand an arborescence in terms of
an adjacency matrix $A\in\mathbb{N}_0^{N\times N}$ containing information about which of the $N$
vertices are connected by which propagators. At this point, it is useful to recall the propagator numbering
based on arrow heads, which has been introduced above: If $A_{ij}=0$, there is no connection from the 
$i^\text{th}$
to the $j^\text{th}$ vertex. Otherwise, it is the $A_{ij}{}^\text{th}$ propagator originating in the
$i^\text{th}$ vertex that establishes this connection. Alternatively, $A_{ij}$ can be understood as the 
number of
arrow heads on the edge directed from the $i^\text{th}$ to the $j^\text{th}$ vertex. Due to orientation, 
each allowed adjacency
matrix is an upper triangular matrix with a vanishing main diagonal. Since all vertices but the
first one have exactly one incoming propagator, there is exactly one non-zero entry in each but
the first column of $A$. 
The set of such triangular matrices over $\mathbb{K}$ is given by
\begin{align}
\mathbb{T}_{\mathbb{K}}^N &= \left\{ \left. M\in \mathbb{K}^{N\times N} \right| \forall i\in\{1,\ldots,N\}\forall j\in\{1,\ldots,i\}:M_{ij}=0 \right. \notag\\
&\hspace{2.8cm}\left. \wedge \left(N>1 \Rightarrow \forall j\in\{2,\ldots,N\}\exists^{=1}i\in\{1,\ldots,N-1\}:M_{ij}\neq 0\right) \right\}.\notag
\end{align}
Then the set of all allowed adjacency matrices for $N$ vertices is the following subset of
$\mathbb{T}_{\mathbb{N}_0}^N$:
\begin{align}
\mathbb{A}^N = &\left\{ \left. A\in\mathbb{T}_{\mathbb{N}_0}^N \right| \forall i\in\{1,\ldots,N-1\}\forall j\in\{2,\ldots,N\}: \left( A_{ij}>0 \Rightarrow \exists j^\prime<j:A_{ij^\prime}=A_{ij}-1 \right)\right\} \label{eq-def-an}
\end{align}
\clearpage
\begin{table}
\begin{minipage}{\textwidth} 
\resizebox{\textwidth}{!}{
\begin{tabular}{||ccc||ccc||}\hhline{|t:===:t:===:t|}
\hspace{0.6cm} MLP-Arborescence \hspace{0.6cm} & $A$ & \hspace{0.6cm}$\alpha(A)$\hspace{0.6cm} & \hspace{0.6cm} MLP-Arborescence \hspace{0.6cm} & $A$ & \hspace{0.6cm} $\alpha(A)$ \hspace{0.35cm}\ \\ \hhline{|:===::===:|}
\begin{tikzpicture}[baseline=-2pt, inner sep=0pt, outer sep=4pt]
    \begin{feynman}
        \vertex [dot] (k1) at (0,0) {};
    \end{feynman}
\end{tikzpicture} & $\begin{pmatrix} 
0
\end{pmatrix}$ & $0\vphantom{\begin{matrix}\ \\ \ \end{matrix}}$ & \begin{tikzpicture}[baseline=-2pt, inner sep=0pt, outer sep=4pt]
    \begin{feynman}
        \vertex [dot] (k1) at (-0.75,0.75) {};
        \vertex [dot] (k2) at (0.75,0.75) {};
				\vertex [dot] (k3) at (0.75,-0.75) {};
        \vertex [dot] (k4) at (-0.75,-0.75) {};
        \diagram*{
				(k1) -- [dn] (k2),
				(k1) -- [d2] (k3),
				(k1) -- [d2] (k4)
        };
    \end{feynman}
\end{tikzpicture} & $\begin{pmatrix} 
0 & 1 & 2 & 2 \\
0 & 0 & 0 & 0 \\
0 & 0 & 0 & 0 \\
0 & 0 & 0 & 0
\end{pmatrix}$ & $2\vphantom{\begin{matrix}\ \\ \ \\ \ \\ \ \\ \ \\ \ \end{matrix}}$ \\ \hhline{|:===:|~~~||}
\begin{tikzpicture}[baseline=-2pt, inner sep=0pt, outer sep=4pt]
    \begin{feynman}
        \vertex [dot] (k1) at (0,0) {};
        \vertex [dot] (k2) at (1.5,0) {};
        \diagram*{
            (k1) -- [dn] (k2)
        };
    \end{feynman}
\end{tikzpicture} & $\begin{pmatrix} 
0 & 1 \\
0 & 0
\end{pmatrix}$ & $1\vphantom{\begin{matrix}\ \\ \ \\ \ \end{matrix}}$  & \begin{tikzpicture}[baseline=-2pt, inner sep=0pt, outer sep=4pt]
    \begin{feynman}
        \vertex [dot] (k1) at (-0.75,0.75) {};
        \vertex [dot] (k2) at (0.75,0.75) {};
				\vertex [dot] (k3) at (0.75,-0.75) {};
        \vertex [dot] (k4) at (-0.75,-0.75) {};
        \diagram*{
				(k1) -- [dn] (k2),
				(k1) -- [dn] (k3),
				(k1) -- [d2] (k4)
        };
    \end{feynman}
\end{tikzpicture} & $\begin{pmatrix} 
0 & 1 & 1 & 2 \\
0 & 0 & 0 & 0 \\
0 & 0 & 0 & 0 \\
0 & 0 & 0 & 0
\end{pmatrix},\begin{pmatrix} 
0 & 1 & 2 & 1 \\
0 & 0 & 0 & 0 \\
0 & 0 & 0 & 0 \\
0 & 0 & 0 & 0
\end{pmatrix}$ & $2\vphantom{\begin{matrix}\ \\ \ \\ \ \\ \ \\ \ \\ \ \end{matrix}}$ \\ \hhline{|:===:|~~~||}
\begin{tikzpicture}[baseline=-2pt, inner sep=0pt, outer sep=4pt]
    \begin{feynman}
        \vertex [dot] (k1) at (0,0) {};
        \vertex [dot] (k2) at (0.87*1.5,0.5*1.5) {};
        \vertex [dot] (k3) at (0.87*1.5,-0.5*1.5) {};
        \diagram*{
            (k1) -- [dn] (k2),
            (k1) -- [d2] (k3)
        };
    \end{feynman}
\end{tikzpicture} & $\begin{pmatrix} 
0 & 1 & 2 \\
0 & 0 & 0 \\
0 & 0 & 0
\end{pmatrix}$ & $2\vphantom{\begin{matrix}\ \\ \ \\ \ \\ \ \\ \ \\ \ \end{matrix}}$  & \begin{tikzpicture}[baseline=-2pt, inner sep=0pt, outer sep=4pt]
    \begin{feynman}
        \vertex [dot] (k1) at (-0.75,0.75) {};
        \vertex [dot] (k2) at (0.75,0.75) {};
				\vertex [dot] (k3) at (0.75,-0.75) {};
        \vertex [dot] (k4) at (-0.75,-0.75) {};
        \diagram*{
				(k2) -- [dn] (k3),
				(k2) -- [d2] (k1) -- [dn] (k4)
        };
    \end{feynman}
\end{tikzpicture} & $\begin{pmatrix} 
0 & 1 & 2 & 0 \\
0 & 0 & 0 & 0 \\
0 & 0 & 0 & 1 \\
0 & 0 & 0 & 0
\end{pmatrix}$ & $2\vphantom{\begin{matrix}\ \\ \ \\ \ \\ \ \\ \ \\ \ \end{matrix}}$ \\ 
\begin{tikzpicture}[baseline=-2pt, inner sep=0pt, outer sep=4pt]
    \begin{feynman}
        \vertex [dot] (k1) at (0,0) {};
        \vertex [dot] (k2) at (0.87*1.5,0.5*1.5) {};
        \vertex [dot] (k3) at (0.87*1.5,-0.5*1.5) {};
        \diagram*{
            (k1) -- [dn] (k2),
            (k1) -- [dn] (k3)
        };
    \end{feynman}
\end{tikzpicture} & $\begin{pmatrix} 
0 & 1 & 1 \\
0 & 0 & 0 \\
0 & 0 & 0
\end{pmatrix}$ & $1\vphantom{\begin{matrix}\ \\ \ \\ \ \\ \ \\ \ \end{matrix}}$  & \begin{tikzpicture}[baseline=-2pt, inner sep=0pt, outer sep=4pt]
    \begin{feynman}
        \vertex [dot] (k1) at (-0.75,0.75) {};
        \vertex [dot] (k2) at (0.75,0.75) {};
				\vertex [dot] (k3) at (0.75,-0.75) {};
        \vertex [dot] (k4) at (-0.75,-0.75) {};
        \diagram*{
				(k1) -- [d2] (k4),
				(k1) -- [dn] (k2) -- [dn] (k3)
        };
    \end{feynman}
\end{tikzpicture} & $\begin{pmatrix} 
0 & 1 & 2 & 0 \\
0 & 0 & 0 & 1 \\
0 & 0 & 0 & 0 \\
0 & 0 & 0 & 0
\end{pmatrix},\begin{pmatrix} 
0 & 1 & 0 & 2 \\
0 & 0 & 1 & 0 \\
0 & 0 & 0 & 0 \\
0 & 0 & 0 & 0
\end{pmatrix}$ & $2\vphantom{\begin{matrix}\ \\ \ \\ \ \\ \ \\ \ \\ \ \end{matrix}}$ \\
\begin{tikzpicture}[baseline=-2pt, inner sep=0pt, outer sep=4pt]
    \begin{feynman}
        \vertex [dot] (k1) at (0,0) {};
        \vertex [dot] (k2) at (0.87*1.5,0.5*1.5) {};
        \vertex [dot] (k3) at (0.87*1.5,-0.5*1.5) {};
        \diagram*{
            (k3) -- [dn] (k1) -- [dn] (k2)
        };
    \end{feynman}
\end{tikzpicture} & $\begin{pmatrix} 
0 & 1 & 0 \\
0 & 0 & 1 \\
0 & 0 & 0
\end{pmatrix}$ & $1\vphantom{\begin{matrix}\ \\ \ \\ \ \\ \ \end{matrix}}$  & \begin{tikzpicture}[baseline=-2pt, inner sep=0pt, outer sep=4pt]
    \begin{feynman}
        \vertex [dot] (k1) at (-0.75,0.75) {};
        \vertex [dot] (k2) at (0.75,0.75) {};
				\vertex [dot] (k3) at (0.75,-0.75) {};
        \vertex [dot] (k4) at (-0.75,-0.75) {};
        \diagram*{
            (k1) -- [dn] (k2),
						(k2) -- [dn] (k3),
						(k1) -- [dn] (k4)
        };
    \end{feynman}
\end{tikzpicture} & $\begin{pmatrix} 
0 & 1 & 1 & 0 \\
0 & 0 & 0 & 1 \\
0 & 0 & 0 & 0 \\
0 & 0 & 0 & 0
\end{pmatrix},\begin{pmatrix} 
0 & 1 & 0 & 1 \\
0 & 0 & 1 & 0 \\
0 & 0 & 0 & 0 \\
0 & 0 & 0 & 0
\end{pmatrix},\begin{pmatrix} 
0 & 1 & 1 & 0 \\
0 & 0 & 0 & 0 \\
0 & 0 & 0 & 1 \\
0 & 0 & 0 & 0
\end{pmatrix}$ & $1\vphantom{\begin{matrix}\ \\ \ \\ \ \\ \ \\ \ \\ \ \end{matrix}}$ \\ \hhline{|:===:|~~~||}
\begin{tikzpicture}[baseline=-2pt, inner sep=0pt, outer sep=4pt]
    \begin{feynman}
        \vertex [dot] (k1) at (-0.75,0.75) {};
        \vertex [dot] (k2) at (0.75,0.75) {};
				\vertex [dot] (k3) at (0.75,-0.75) {};
        \vertex [dot] (k4) at (-0.75,-0.75) {};
        \diagram*{
				(k1) -- [dn] (k2),
				(k1) -- [d2] (k3),
				(k1) -- [d3] (k4)
        };
    \end{feynman}
\end{tikzpicture} & $\begin{pmatrix} 
0 & 1 & 2 & 3 \\
0 & 0 & 0 & 0 \\
0 & 0 & 0 & 0 \\
0 & 0 & 0 & 0
\end{pmatrix}$ & $3\vphantom{\begin{matrix}\ \\ \ \\ \ \\ \ \\ \ \\ \ \end{matrix}}$  & \begin{tikzpicture}[baseline=-2pt, inner sep=0pt, outer sep=4pt]
    \begin{feynman}
        \vertex [dot] (k1) at (-0.75,0.75) {};
        \vertex [dot] (k2) at (0.75,0.75) {};
				\vertex [dot] (k3) at (0.75,-0.75) {};
        \vertex [dot] (k4) at (-0.75,-0.75) {};
        \diagram*{
            (k4) -- [dn] (k1) -- [d2] (k2),
            (k1) -- [dn] (k3)
        };
    \end{feynman}
\end{tikzpicture} & $\begin{pmatrix} 
0 & 1 & 0 & 0 \\
0 & 0 & 1 & 2 \\
0 & 0 & 0 & 0 \\
0 & 0 & 0 & 0
\end{pmatrix}$ & $2\vphantom{\begin{matrix}\ \\ \ \\ \ \\ \ \\ \ \\ \ \end{matrix}}$ \\
\begin{tikzpicture}[baseline=-2pt, inner sep=0pt, outer sep=4pt]
    \begin{feynman}
        \vertex [dot] (k1) at (-0.75,0.75) {};
        \vertex [dot] (k2) at (0.75,0.75) {};
				\vertex [dot] (k3) at (0.75,-0.75) {};
        \vertex [dot] (k4) at (-0.75,-0.75) {};
        \diagram*{
				(k1) -- [dn] (k2),
				(k1) -- [dn] (k3),
				(k1) -- [dn] (k4)
        };
    \end{feynman}
\end{tikzpicture} & $\begin{pmatrix} 
0 & 1 & 1 & 1 \\
0 & 0 & 0 & 0 \\
0 & 0 & 0 & 0 \\
0 & 0 & 0 & 0
\end{pmatrix}$ & $1\vphantom{\begin{matrix}\ \\ \ \\ \ \\ \ \\ \ \\ \ \end{matrix}}$  & \begin{tikzpicture}[baseline=-2pt, inner sep=0pt, outer sep=4pt]
    \begin{feynman}
        \vertex [dot] (k1) at (-0.75,0.75) {};
        \vertex [dot] (k2) at (0.75,0.75) {};
				\vertex [dot] (k3) at (0.75,-0.75) {};
        \vertex [dot] (k4) at (-0.75,-0.75) {};
        \diagram*{
            (k4) -- [dn] (k1) -- [dn] (k2),
            (k1) -- [dn] (k3)
        };
    \end{feynman}
\end{tikzpicture} & $\begin{pmatrix} 
0 & 1 & 0 & 0 \\
0 & 0 & 1 & 1 \\
0 & 0 & 0 & 0 \\
0 & 0 & 0 & 0
\end{pmatrix}$ & $1\vphantom{\begin{matrix}\ \\ \ \\ \ \\ \ \\ \ \\ \ \end{matrix}}$ \\ \hhline{||~~~|:===:b|}
\begin{tikzpicture}[baseline=-2pt, inner sep=0pt, outer sep=4pt]
    \begin{feynman}
        \vertex [dot] (k1) at (-0.75,0.75) {};
        \vertex [dot] (k2) at (0.75,0.75) {};
				\vertex [dot] (k3) at (0.75,-0.75) {};
        \vertex [dot] (k4) at (-0.75,-0.75) {};
        \diagram*{
            (k4) -- [dn] (k1) -- [dn] (k2) -- [dn] (k3)
        };
    \end{feynman}
\end{tikzpicture} & $\begin{pmatrix} 
0 & 1 & 0 & 0 \\
0 & 0 & 1 & 0 \\
0 & 0 & 0 & 1 \\
0 & 0 & 0 & 0
\end{pmatrix}$ & $1\vphantom{\begin{matrix}\ \\ \ \\ \ \\ \ \end{matrix}}$  &  \multicolumn{3}{c}{\ }

\\ \hhline{|b:===:b|~~~:|}
\end{tabular}
}
\end{minipage}
\caption{All arborescences and corresponding adjacency matrices $A\in\mathbb{A}^N$ as well as
saturation thresholds $\alpha(A)$ for $N=1,2,3,4$ vertices. In the way arborescences are represented
here, it is always the vertex $k_1$ that serves as root. From here, we can deduce that the maximum
saturation threshold among all arborescences with $N$ vertices is $N-1$.}
\label{tab-arbor-ex}
\end{table}
Counting the appearances of $A_{ij}>0$ in the $i^\text{th}$ line of a given adjacency matrix
$A\in\mathbb{A}^N$ determines the order of the respective propagator: If $A_{ij}$ appears $n-1$
times in the $i^\text{th}$ line, the corresponding propagator is of order $n$. Note that the
appearance of a propagator of order $n$ decreases the saturation threshold $\alpha(A)$ by $n-2$.
Another source that leads to smaller $\alpha(A)$ are internal vertices: $\alpha(A)$ is
decreased by $1$ for each internal vertex, or, in terms of adjacency matrices, for each non-zero
line but the first one. This behavior is entirely described by the simple expression
\begin{equation}
\alpha(A)=\max\limits_{i,j} A_{ij}. \label{eq-alpha-def}
\end{equation}
If $l$ reaches or undershoots $\alpha(A)$, the corresponding arborescence is saturated and
does not contribute to $\Delta_{mk_1\ldots k_{N}}^{(l,N)}$. Tab.~\ref{tab-arbor-ex} lists example
arborescences and corresponding adjacency matrices $A$ as well as saturation thresholds $\alpha(A)$.
Note that there may be several allowed adjacency matrices for one arborescence due to index permutation symmetry.
The only thing left for expressing $\Delta_{mk_1\ldots k_{N}}^{(l,N)}$ solely in terms of propagators and
biases is an analytical representation $\delta_{mk_1\ldots k_N}^{(l,N)}(A)$ of an individual arborescence
with $N$ vertices for a given adjacency matrix $A$. For its formulation, we use the function
\begin{equation}
\beta_{c}(A) = \sum\limits_{i=1}^N i\cdot \Theta(A_{ic}) \label{eq-beta-def}
\end{equation}
that determines the line in which the entry in the $j^\text{th}$ column is non-zero. That means it provides the unique vertex, from which a propagator leads to the $c^\text{th}$ vertex. Since there
is no antecedent propagator to the root of an arborescence, Eq.~\eqref{eq-beta-def} vanishes for $c=1$. 
Using Eq.~\eqref{eq-beta-def}, the abbreviations
\begin{equation}
j_c(A)=j_{A_{\beta_c(A)\ c}}^{\{\beta_c(A)\}}, \hspace{0.5cm} q_c(A)=q_{j_c(A)}^{\beta_c(A)}, \hspace{0.5cm} n_{bc}(A)=\sum\limits_{j=1}^N \delta_{b\ A_{cj}}, \hspace{0.5cm} D_{bc}^{(p)}(A) = D_{q^{\{c\}}_{j_b^{\{c\}}+1}q^{\{c\}}_{j_b^{\{c\}}}}^{\left(j_b^{\{c\}},p\right)}, \notag
\end{equation} 
and the previous observations, we write
\begin{align}
\delta_{mk_1\ldots k_N}^{(l,N)}(A) = \sum\limits_{q^{\{ 0\}}_l}\delta_{mq_l^{\{ 0\}}}&\sum\limits_{j_{A_{01}}^{\{0\}}}\delta_{lj_{A_{01}}^{\{0\}}}\prod\limits_{c=1}^N \mlpsum\limits_{(q_a^{\{c\}})_{a=1}^{j_c(A)} \vphantom{\begin{matrix}{}\\{}\end{matrix}}\ (j_b^{\{c\}})_{b=1}^{\mathrm{max}_{r}(A_{cr})}}\hspace{-1.5cm}{\vphantom{\mlpsum}}_{q_c(A)\ k_c}\prod\limits_{b=1}^{\mathrm{max}_{r}(A_{cr})} D_{bc}^{(1+n_{bc}(A))}(A).\label{eq-smalldelta-def}
\end{align}

%
Eq.~\eqref{eq-smalldelta-def} may appear very intimidating at first sight, but its individual
terms are easy to interpret and to recognize in the summands of Eqs.~\eqref{eq-vertex-1},
\eqref{eq-deltak1k2-x1} and \eqref{eq-deltak1k2k3-x1}:
\begin{itemize}
\item The expression $\mathrm{max}_{r}(A_{cr})$ corresponds to the number of all propagators
  originating in the $c^\text{th}$ vertex. Thus, all these propagators are collected by the product
	\begin{equation*} \prod\limits_{b=1}^{\mathrm{max}_{r}(A_{cr})} D_{bc}^{(1+n_{bc}(A))}(A). \end{equation*}
	The order of the $b^\text{th}$ propagator originating in the $c^\text{th}$ vertex is equal
        to $1$ plus the total number $n_{bc}(A)$ of appearances of the entry $b$ in the $c^\text{th}$ line of
        $A$. If the $c^\text{th}$ vertex is a leaf of the arborescence, that is if it is external
        such that there are no outgoing propagators and $\mathrm{max}_{r}(A_{cr})=0$, the product
        can be neglected.
	\item The $c^\text{th}$ vertex of the arborescence is denoted by the expression
	\begin{equation*} \mlpsum\limits_{(q_a^{\{c\}})_{a=1}^{j_c(A)} \vphantom{\begin{matrix}{}\\{}\end{matrix}}\ (j_b^{\{c\}})_{b=1}^{\mathrm{max}_{r}(A_{cr})}}\hspace{-1.5cm}{\vphantom{\mlpsum}}_{q_c(A)\ k_c}.\end{equation*}
	\vspace{-1cm}
	
	\noindent
	For each of the $\mathrm{max}_{r}(A_{cr})$ propagators that originate in the $c^\text{th}$
        vertex, there is a summation index $j_b^{\{c\}}$. It is the $\beta_c(A)$-th vertex, whose
        $A_{\beta_c(A)c}$-th propagator leads to the $c^\text{th}$ vertex. This is the reason, why we
        sum over the $(q_c(A),k_c)$-th matrix element of the
        $c^\text{th}$ vertex in the $j_c(A)$-th layer.
	\item All $N$ vertices of the entire arborescence are collected by
\begin{equation*} \sum\limits_{q^{\{ 0\}}_l}\delta_{mq_l^{\{ 0\}}}\sum\limits_{j_{A_{01}}^{\{0\}}}\delta_{lj_{A_{01}}^{\{0\}}}\prod\limits_{c=1}^N. \end{equation*}
\end{itemize} 
\vfill
\noindent
Using Eq.~\eqref{eq-smalldelta-def} and taking the corresponding saturation thresholds given in
Eq.~\eqref{eq-alpha-def} into account, we can represent $\Delta_{mk_1\ldots k_N}^{(l,N)}$ as the
following sum over all adjacency matrices $A$ in $\mathbb{A}^N$ (see App.~A on
p.~\pageref{sec-proofs-2}):
\vfill
\noindent
\begin{equation}
  \Delta_{mk_1\ldots k_N}^{(l,N)} = \sum\limits_{A\in \mathbb{A}^N} \Theta(l-\alpha(A))\
  \delta_{mk_1\ldots k_N}^{(l,N)}(A). \label{eq-delta-sum}
\end{equation}
\vfill
\noindent
Finally, Eq.~\eqref{eq-delta-sum} can be inserted into Eq.~\eqref{eq-propderiv-2}, which is
required for computing the Taylor coefficients of~$\bm{\mathcal{Y}}$ as shown in Eq.~\eqref{eq-yderiv-1}.
Considering
Eq.~\eqref{eq-propderiv-2}, it is important to note that none of the indices $k_1,\ldots,k_N$
is shared among several factors $\Delta^{(l,\pi_i)}$. Therefore, each summand of $\partial^N
D_{nm}^{(l,p)}/(\partial x_{k_1}\ldots \partial x_{k_N})$ consists of arborescences that each contain
different vertices. The effective saturation threshold of a product of several $\Delta^{(l,\pi_i)}$
is, thereby, simply the maximal saturation threshold among all given factors.
The Taylor coefficients up to third order then turn out as
\vfill
\noindent
\begin{align}
\left. \frac{\partial \mathcal{Y}_n}{\partial x_{k_1}}\right|_{\bm{x}=\bm{x}_0}&=\sum\limits_{m=1}^{H_{L-1}} D_{nm}^{(L-1,1)} \Theta(L-1)\left. \left(\ \begin{tikzpicture}[baseline=-2pt, inner sep=0pt, outer sep=4pt]
    \begin{feynman}
        \vertex [dot] (k1) at (0,0) {};
        \diagram*{
            (k1) -- (k1)
        };
    \end{feynman}
\end{tikzpicture}\vphantom{\sum}\ \right)_{mk_1}^{(L-1,1)} \right|_{\bm{x}=\bm{x}_0} \label{eq-taylor1-x2}
\\
\left. \frac{\partial^2 \mathcal{Y}_n}{\partial x_{k_1}\ \partial x_{k_2}}\right|_{\bm{x}=\bm{x}_0}&=\sum\limits_{m=1}^{H_{L-1}}\left[ D_{nm}^{(L-1,1)}\Theta(L-2)\left( \begin{tikzpicture}[baseline=-2pt, inner sep=0pt, outer sep=4pt]
    \begin{feynman}
        \vertex [dot] (k1) at (0,0) {};
        \vertex [dot] (k2) at (1,0) {};
        \diagram*{
            (k1) -- [dn] (k2)
        };
    \end{feynman}
\end{tikzpicture}\vphantom{\sum}\right)_{mk_1k_2}^{(L-1,2)} \right. \notag \\ 
&\ \hspace{0.9cm} + \left. \left. D_{nm}^{(L-1,2)}\Theta(L-1)\left(\ \begin{tikzpicture}[baseline=-2pt, inner sep=0pt, outer sep=4pt]
    \begin{feynman}
        \vertex [dot] (k1) at (0,0) {};
        \diagram*{
            (k1) -- (k1)
        };
    \end{feynman}
\end{tikzpicture}\vphantom{\sum}\ \right)_{mk_1}^{(L-1,1)}\left(\ \begin{tikzpicture}[baseline=-2pt, inner sep=0pt, outer sep=4pt]
    \begin{feynman}
        \vertex [dot] (k1) at (0,0) {};
        \diagram*{
            (k1) -- (k1)
        };
    \end{feynman}
\end{tikzpicture}\vphantom{\sum}\ \right)_{mk_2}^{(L-1,1)} \right] \right|_{\bm{x}=\bm{x}_0} \label{eq-taylor2-x2}
\\ 
\left. \frac{\partial^3 \mathcal{Y}_n}{\partial x_{k_1}\ \partial x_{k_2}\ \partial x_{k_3}}\right|_{\bm{x}=\bm{x}_0}&=\sum\limits_{m=1}^{H_{L-1}} \scalebox{0.89}{$ \left\{ D_{nm}^{(L-1,1)}  \left[ \Theta(L-3) \left( \begin{tikzpicture}[baseline=-2pt, inner sep=0pt, outer sep=4pt]
    \begin{feynman}
        \vertex [dot] (k1) at (0,0) {};
        \vertex [dot] (k2) at (0.87,0.5) {};
        \vertex [dot] (k3) at (0.87,-0.5) {};
        \diagram*{
            (k1) -- [dn] (k2),
            (k1) -- [d2] (k3)
        };
    \end{feynman}
\end{tikzpicture}\ \right)_{mk_1k_2k_3}^{(L-1,3)} +\Theta(L-2) \left( \begin{tikzpicture}[baseline=-2pt, inner sep=0pt, outer sep=4pt]
    \begin{feynman}
        \vertex [dot] (k1) at (0,0) {};
        \vertex [dot] (k2) at (0.87,0.5) {};
        \vertex [dot] (k3) at (0.87,-0.5) {};
        \diagram*{
            (k1) -- [dn] (k2),
            (k1) -- [dn] (k3)
        };
    \end{feynman}
\end{tikzpicture}\ \right)_{mk_1k_2k_3}^{(L-1,3)} \right. \right. $} \notag 
\\ 
&\ \hspace{0.95cm} \scalebox{0.89}{$+ \left. \Theta(L-2) \left( \begin{tikzpicture}[baseline=-2pt, inner sep=0pt, outer sep=4pt]
    \begin{feynman}
        \vertex [dot] (k1) at (0,0) {};
        \vertex [dot] (k2) at (0.87,0.5) {};
        \vertex [dot] (k3) at (0.87,-0.5) {};
        \diagram*{
            (k3) -- [dn] (k1) -- [dn] (k2)
        };
    \end{feynman}
\end{tikzpicture}\ \right)_{mk_1k_2k_3}^{(L-1,3)} \right] + D_{nm}^{(L-1,2)} \left[ \Theta(L-2) \left(\ \begin{tikzpicture}[baseline=-2pt, inner sep=0pt, outer sep=4pt]
    \begin{feynman}
        \vertex [dot] (k1) at (0,0) {};
        \diagram*{
            (k1) -- (k1)
        };
    \end{feynman}
\end{tikzpicture}\vphantom{\sum}\ \right)_{mk_1}^{(L-1,1)}\left(\ \begin{tikzpicture}[baseline=-2pt, inner sep=0pt, outer sep=4pt]
    \begin{feynman}
        \vertex [dot] (k1) at (0,0.5) {};
				\vertex [dot] (k2) at (0,-0.5) {};
        \diagram*{
            (k1) -- [dn] (k2)
        };
    \end{feynman}
\end{tikzpicture}\vphantom{\sum}\ \right)_{mk_2k_3}^{(L-1,2)} \right. $} \notag 
\\ 
&\ \hspace{0.94cm} \scalebox{0.9}{$+\ \Theta(L-2) \left(\ \begin{tikzpicture}[baseline=-2pt, inner sep=0pt, outer sep=4pt]
    \begin{feynman}
        \vertex [dot] (k1) at (0,0) {};
        \diagram*{
            (k1) -- (k1)
        };
    \end{feynman}
\end{tikzpicture}\vphantom{\sum}\ \right)_{mk_2}^{(L-1,1)}\left(\ \begin{tikzpicture}[baseline=-2pt, inner sep=0pt, outer sep=4pt]
    \begin{feynman}
        \vertex [dot] (k1) at (0,0.5) {};
				\vertex [dot] (k2) at (0,-0.5) {};
        \diagram*{
            (k1) -- [dn] (k2)
        };
    \end{feynman}
\end{tikzpicture}\vphantom{\sum}\ \right)_{mk_1k_3}^{(L-1,2)} + \left. \Theta(L-2) \left(\ \begin{tikzpicture}[baseline=-2pt, inner sep=0pt, outer sep=4pt]
    \begin{feynman}
        \vertex [dot] (k1) at (0,0) {};
        \diagram*{
            (k1) -- (k1)
        };
    \end{feynman}
\end{tikzpicture}\vphantom{\sum}\ \right)_{mk_3}^{(L-1,1)}\left(\ \begin{tikzpicture}[baseline=-2pt, inner sep=0pt, outer sep=4pt]
    \begin{feynman}
        \vertex [dot] (k1) at (0,0.5) {};
				\vertex [dot] (k2) at (0,-0.5) {};
        \diagram*{
            (k1) -- [dn] (k2)
        };
    \end{feynman}
\end{tikzpicture}\vphantom{\sum}\ \right)_{mk_1k_2}^{(L-1,2)} \right] $} \notag 
\\ 
&\  \hspace{0.94cm} \scalebox{0.9}{$+\ D_{nm}^{(L-1,3)}\ \ \, \Theta(L-1)\ \ \, \left. \left. \left(\ \begin{tikzpicture}[baseline=-2pt, inner sep=0pt, outer sep=4pt]
    \begin{feynman}
        \vertex [dot] (k1) at (0,0) {};
        \diagram*{
            (k1) -- (k1)
        };
    \end{feynman}
\end{tikzpicture}\vphantom{\sum}\ \right)_{mk_1}^{(L-1,1)}\left(\ \begin{tikzpicture}[baseline=-2pt, inner sep=0pt, outer sep=4pt]
    \begin{feynman}
        \vertex [dot] (k1) at (0,0) {};
        \diagram*{
            (k1) -- (k1)
        };
    \end{feynman}
\end{tikzpicture}\vphantom{\sum}\ \right)_{mk_2}^{(L-1,1)}\left(\ \begin{tikzpicture}[baseline=-2pt, inner sep=0pt, outer sep=4pt]
    \begin{feynman}
        \vertex [dot] (k1) at (0,0) {};
        \diagram*{
            (k1) -- (k1)
        };
    \end{feynman}
\end{tikzpicture}\vphantom{\sum}\ \right)_{mk_3}^{(L-1,1)}\vphantom{\vphantom{\begin{matrix}a\\a\\a\end{matrix}}} \right\} \right|_{\bm{x}=\bm{x}_0} $}  \label{eq-taylor3-x2}
\end{align}
\vfill
\noindent
Let $I_j=\{i^{j}_{1},\ldots,i^{j}_{y_j}\}\subseteq \{1,\ldots,N\}$ with $j\in\{1,\ldots,p\}$ be pairwise disjoint index sets such that ${I_1\cup\ldots\cup I_p = \{1,\ldots,N\}}$ and $I_{j_1}\cap I_{j_2} = \emptyset$ for $j_1\neq j_2$. This implies $y_1+\ldots+y_p=N$. Given these indices, the following short-hand notation proves useful for the Taylor series, since it helps to combine $p$ arborescences covering $N$ vertices to one disconnected graph with $p$ connected components:\clearpage
\begin{align}
&\ \hspace{4cm}\left( \begin{tikzpicture}[baseline=-2pt, inner sep=0pt, outer sep=4pt]
    \begin{feynman}
		\vertex [dot] (k1) at (-1,1) {};
		\vertex [dot] (k2) at (-1,-1) {};
		\vertex [dot] (kmid) at (-1,0) {};
		\coordinate (u1) at (-2,1);
		\coordinate (umid) at (-2,0);
		\coordinate (u2) at (-2,-1);
		\coordinate (mid) at (0,0) {};
        \diagram*{
				mid [blob],
				(mid) -- (k1),
				(mid) -- (kmid),
				(mid) -- (k2),
				(u1) -- [photon] (k1),
				(umid) -- [photon] (kmid),
				(u2) -- [photon] (k2)
				};
    \end{feynman}
		\node at (-1,0.59) {$\vdots$};
		\node at (-1,-0.41) {$\vdots$};
\end{tikzpicture} \ \ldots\ \begin{tikzpicture}[baseline=-2pt, inner sep=0pt, outer sep=4pt]
    \begin{feynman}
		\vertex [dot] (k1) at (1,1) {};
		\vertex [dot] (k2) at (1,-1) {};
		\vertex [dot] (kmid) at (1,0) {};
		\coordinate (u1) at (2,1);
		\coordinate (umid) at (2,0);
		\coordinate (u2) at (2,-1);
		\coordinate (mid) at (0,0) {};
        \diagram*{
				mid [blob],
				(mid) -- (k1),
				(mid) -- (kmid),
				(mid) -- (k2),
				(u1) -- [photon] (k1),
				(umid) -- [photon] (kmid),
				(u2) -- [photon] (k2)
				};
    \end{feynman}
		\node at (1,0.59) {$\vdots$};
		\node at (1,-0.41) {$\vdots$};
\end{tikzpicture}\right)_{n}^{\bm{x},\bm{x}_0} \notag \\&=
\sum\limits_{m=1}^{H_{L-1}} D_{nm}^{(L-1,p)}\sum\limits_{k_1}^{H_0}\ldots\sum\limits_{k_N}^{H_0} \left. \left( \begin{tikzpicture}[baseline=-2pt, inner sep=0pt, outer sep=4pt]
    \begin{feynman}
		\vertex [dot] (k1) at (-1,1) {};
		\vertex [dot] (k2) at (-1,-1) {};
		\vertex [dot] (kmid) at (-1,0) {};
		\coordinate  (dots1) at (-1,0.5) {};
		\coordinate (mid) at (0,0) {};
        \diagram*{
				mid [blob],
				(mid) -- (k1),
				(mid) -- (kmid),
				(mid) -- (k2)
				};
    \end{feynman}
		\node at (-1,0.59) {$\vdots$};
		\node at (-1,-0.41) {$\vdots$};
\end{tikzpicture}\right)_{mk_{i^{1}_1}\ldots k_{i^{1}_{y_1}}}^{(L-1,y_1)}\ldots \left( \begin{tikzpicture}[baseline=-2pt, inner sep=0pt, outer sep=4pt]
    \begin{feynman}
		\vertex [dot] (k1) at (1,1) {};
		\vertex [dot] (k2) at (1,-1) {};
		\vertex [dot] (kmid) at (1,0) {};
		\coordinate  (dots1) at (1,0.5) {};
		\coordinate (mid) at (0,0) {};
        \diagram*{
				mid [blob],
				(mid) -- (k1),
				(mid) -- (kmid),
				(mid) -- (k2)
				};
    \end{feynman}
		\node at (1,0.59) {$\vdots$};
		\node at (1,-0.41) {$\vdots$};
\end{tikzpicture}\right)_{mk_{i^{p}_1}\ldots k_{i^{p}_{y_1}}}^{(L-1,y_p)}\ \right|_{\bm{x}=\bm{x}_0} \prod\limits_{i=1}^N(\bm{x}-\bm{x}_0)_{k_i}. \label{eq-contraction}
\end{align}
\vfill
\noindent
The resulting graph is only connected for $p=1$. Using the Taylor coefficients 
from Eqs.~\eqref{eq-taylor1-x2} to \eqref{eq-taylor3-x2} and the short-hand notation 
from Eq.~\eqref{eq-contraction} finally yields the
interesting series representation, which is ordered by the number of connected components:
\vfill
\noindent
\vfill
\noindent
\begin{align}
\bm{\mathcal{Y}}(\bm{x})-\bm{\mathcal{Y}}(\bm{x}_0) = &\ \ \ \ \ \Theta(L-1)\left(\ \begin{tikzpicture}[baseline=-2pt, inner sep=0pt, outer sep=4pt]
    \begin{feynman}
        \vertex [dot] (k1) at (0,0) {};
				\coordinate (u1) at (-1,0);
        \diagram*{
            (u1) -- [photon] (k1)
        };
    \end{feynman}
\end{tikzpicture}\vphantom{\sum}\ \right)^{\bm{x},\bm{x}_0}+\frac{\Theta(L-2)}{2}\left( \begin{tikzpicture}[baseline=-2pt, inner sep=0pt, outer sep=4pt]
    \begin{feynman}
        \vertex [dot] (k1) at (0,0) {};
        \vertex [dot] (k2) at (1,0) {};
				\coordinate (u1) at (-1,0);
				\coordinate (u2) at (2,0);
        \diagram*{
            (u1) -- [photon] (k1) -- [dn] (k2) -- [photon] (u2)
        };
    \end{feynman}
\end{tikzpicture}\vphantom{\sum}\right)^{\bm{x},\bm{x}_0}\notag\\
&+\frac{\Theta(L-3)}{6}\left( \begin{tikzpicture}[baseline=-2pt, inner sep=0pt, outer sep=4pt]
    \begin{feynman}
        \vertex [dot] (k1) at (0,0) {};
        \vertex [dot] (k2) at (0.87,0.5) {};
        \vertex [dot] (k3) at (0.87,-0.5) {};
				\coordinate (u1) at (-1,0);
				\coordinate (u2) at (1.87,0.5);
				\coordinate (u3) at (1.87,-0.5);
        \diagram*{
            (k1) -- [dn] (k2),
            (k1) -- [d2] (k3),
						(u1) -- [photon] (k1),
						(u2) -- [photon] (k2),
						(u3) -- [photon] (k3)
        };
    \end{feynman}
\end{tikzpicture}\ \right)^{\bm{x},\bm{x}_0}+\frac{\Theta(L-2)}{6}\left( \begin{tikzpicture}[baseline=-2pt, inner sep=0pt, outer sep=4pt]
    \begin{feynman}
        \vertex [dot] (k1) at (0,0) {};
        \vertex [dot] (k2) at (0.87,0.5) {};
        \vertex [dot] (k3) at (0.87,-0.5) {};
				\coordinate (u1) at (-1,0);
				\coordinate (u2) at (1.87,0.5);
				\coordinate (u3) at (1.87,-0.5);
        \diagram*{
            (k1) -- [dn] (k2),
            (k1) -- [dn] (k3),
						(u1) -- [photon] (k1),
						(u2) -- [photon] (k2),
						(u3) -- [photon] (k3)
        };
    \end{feynman}
\end{tikzpicture}\ \right)^{\bm{x},\bm{x}_0} \notag \\
&+ \frac{\Theta(L-2)}{6}\left( \begin{tikzpicture}[baseline=-2pt, inner sep=0pt, outer sep=4pt]
    \begin{feynman}
        \vertex [dot] (k1) at (0,0) {};
        \vertex [dot] (k2) at (0.87,0.5) {};
        \vertex [dot] (k3) at (0.87,-0.5) {};
				\coordinate (u1) at (-1,0);
				\coordinate (u2) at (1.87,0.5);
				\coordinate (u3) at (1.87,-0.5);
        \diagram*{
            (k1) -- [dn] (k2),
            (k3) -- [dn] (k1),
						(u1) -- [photon] (k1),
						(u2) -- [photon] (k2),
						(u3) -- [photon] (k3)
        };
    \end{feynman}
\end{tikzpicture}\ \right)^{\bm{x},\bm{x}_0} + \ldots \notag \\ &\ \notag \\
&+\frac{\Theta(L-1)}{2}\left( \begin{tikzpicture}[baseline=-2pt, inner sep=0pt, outer sep=4pt]
    \begin{feynman}
        \vertex [dot] (k1) at (0,0) {};
        \vertex [dot] (k2) at (1,0) {};
				\coordinate (u1) at (-1,0);
				\coordinate (u2) at (2,0);
        \diagram*{
            (u1) -- [photon] (k1),
						(k2) -- [photon] (u2)
        };
    \end{feynman}
\end{tikzpicture}\vphantom{\sum}\right)^{\bm{x},\bm{x}_0} + \frac{\Theta(L-2)}{2}\left( \begin{tikzpicture}[baseline=-2pt, inner sep=0pt, outer sep=4pt]
    \begin{feynman}
        \vertex [dot] (k1) at (0,0) {};
        \vertex [dot] (k2) at (0.87,0.5) {};
        \vertex [dot] (k3) at (0.87,-0.5) {};
				\coordinate (u1) at (-1,0);
				\coordinate (u2) at (1.87,0.5);
				\coordinate (u3) at (1.87,-0.5);
        \diagram*{
            (k2) -- [dn] (k3),
						(u1) -- [photon] (k1),
						(u2) -- [photon] (k2),
						(u3) -- [photon] (k3)
        };
    \end{feynman}
\end{tikzpicture}\ \right)^{\bm{x},\bm{x}_0}+\ldots \notag \\ &\ \notag \\
&+\frac{\Theta(L-2)}{2}\left( \begin{tikzpicture}[baseline=-2pt, inner sep=0pt, outer sep=4pt]
    \begin{feynman}
        \vertex [dot] (k1) at (0,0) {};
        \vertex [dot] (k2) at (0.87,0.5) {};
        \vertex [dot] (k3) at (0.87,-0.5) {};
				\coordinate (u1) at (-1,0);
				\coordinate (u2) at (1.87,0.5);
				\coordinate (u3) at (1.87,-0.5);
        \diagram*{
            (u1) -- [photon] (k1),
						(u2) -- [photon] (k2),
						(u3) -- [photon] (k3)
        };
    \end{feynman}
\end{tikzpicture}\ \right)^{\bm{x},\bm{x}_0} + \ldots \notag \\ &\ \notag \\
&+ \ldots
\end{align}
\vfill
\noindent
Graphs that contain up to $N$ vertices contribute to the $N^\text{th}$ Taylor approximation 
of $\bm{\mathcal{Y}}$. 
The deeper $\bm{\mathcal{Y}}$, that is the larger $L$, the more graphs contribute to a given 
order of the Taylor 
expansion due to overshooting the corresponding saturation threshold.\clearpage

\section{NN Perturbation Theory applied to Scattering Lengths}
\label{sec:network}

Depending on the analytical structure of the target function, there may be a difference of several 
orders of magnitude between the Taylor coefficients of different orders. For example, if the 
first-order Taylor coefficients dominate and if all training samples are closely distributed around 
the expansion point, then a trained MLP basically applies a linear approximation to imitate the 
target function. As a supervisor one does not gain any insights on higher-order terms in such cases, 
since these presumably are not faithful to the derivatives of the target function. As can be seen 
in Eq.~\eqref{eq-born-kernel-x3}, the first and second-order Taylor coefficients of the sampled 
Born series vary by a factor $1/(H_0^2)\approx 10^{-3}$. The sampling rate $H_0$ must be sufficiently 
large such that the discretization error becomes negligible, which we assume to be the case for 
$H_0=32$, as used in the following analysis. For the specific application, this implies that we cannot 
expect to recover both, the first and second-order Born terms, from a naively trained MLP or ensemble. 
Therefore, we propose an iterative scheme to gain information on Taylor coefficients of successively 
rising order. Given a training set $T_1^{(0)}$ and test set $T_2^{(0)}$ and assuming we do not train 
single MLPs, but ensembles of several MLPs, the $i^\text{th}$ iteration contains the following steps:
\begin{enumerate}
	\item Initialize a new ensemble $\bm{\mathcal{Y}}^{(i)}=\{\bm{\mathcal{Y}}_n^{(i)}\}_{n=1}^{N_i}$ 
	of MLPs $\bm{\mathcal{Y}}_n^{(i)}$. The output of the ensemble is simply the mean of the individual 
	member outputs.
	\item Train the ensemble $\bm{\mathcal{Y}}^{(i)}$ on the training set $T_1^{(i-1)}$ and validate it 
	later using the test set $T_2^{(i-1)}$.
	\item Compute the $i^\text{th}$-order term $1/(N_ii!)\sum_n\sum_{k_1}\ldots\sum_{k_i} (\partial^i \bm{\mathcal{Y}}_n^{(i)})/(\partial x_{k_1}\ldots \partial x_{k_i}) |_{\bm{x}=\bm{x}_0} \prod_j (\bm{x}-\bm{x}_0)_{k_j} $ of the Taylor expansion of 
	$\sum_n\bm{\mathcal{Y}}_n^{(i)}/N_i$ around the expansion point $\bm{x}_0$. As this term is of leading 
	order, the corresponding $i^\text{th}$-order derivatives and, therefore, Taylor 
	coefficients can be assumed to be faithful to the analytical structure of the target function.
	\item Generate new training and test sets $T_1^{(i)}$ and $T_2^{(i)}$ by substracting the leading 
	order term of the previous step from the targets of $T_1^{(i-1)}$ and $T_2^{(i-1)}$, respectively. 
	If necessary, the targets must be normalized or standardized again.
\end{enumerate}
At the cost of rerunning the training pipeline for each iteration anew, we especially expect this 
procedure to yield faithful first- and second-order Taylor coefficients in the case of S-wave 
scattering lengths. Note that such an iterative approach is anything but unnatural and is really 
just the central idea of perturbation theory.

At first we generate a training and test set of shallow potentials without any bound states. The scattering 
lengths for sampled potentials are derived using the Transfer Matrix Method, see Ref.~\cite{Jonsson:1990}, 
and are uniformly distributed between the boundaries $a_0=-1$ and $a_0= 0$. The training and test set 
contain $|T_1^{(0)}|=3\times 10^4$ and $|T_2^{(0)}|=3\times 10^3$ samples, respectively.

Training and validation of the ensembles at each iteration is performed in PyTorch \cite{Paszke:2019}. At 
first, we initialize an ensemble $\bm{\mathcal{Y}}^{(1)}$ of $N_1=10^2$ MLPs, in which each but the output 
layer is activated via the GELU activation function \cite{Hendrycks:2020}. GELU is smooth in the origin 
in contrast to other rectifiers like ReLU. Being a rectifier, it bypasses the vanishing-gradients-problem, which makes it 
particularly
interesting for deeper architectures, see Ref.~\cite{Liu:2019lzgql}. Their weights and biases are initialized using 
He-Initialization~\cite{He:2015hzrs}. Apart from the requirement of being a GELU-activated MLP, we allow 
various numbers of layers $L_n^{(i)}$, numbers $(H_l)_n^{(i)}$ of units per hidden layer, learning rates 
$\eta_n^{(i)}$ and weight decays $\lambda_n^{(i)}$: The former two are uniformly distributed random 
integers in the intervals $[3, 10]$ and $[16, 256]$, respectively. For the random floats 
$\overline{\eta}_n^{(1)}$ and $\overline{\lambda}_n^{(1)}$ drawn from the uniform distributions over the 
intervals $[2, 3]$ and $[3, 5]$, we work with an exponentially decaying learning rate schedule
\begin{equation}
(\eta_n^{(i)})_\epsilon = \exp\left(-\frac{\epsilon-1}{ \overline{\eta}_n^{(i)}}\right) \cdot 10^{-\overline{\eta}_n^{(i)}}
\end{equation}
\vspace{-0.5cm}
and with the weight-decay
\begin{equation}
\lambda_n^{(i)} = 10^{-\overline{\lambda}_n^{(i)}}.
\end{equation}
The index $\epsilon$ labels the current training epoch and ranges from $\epsilon=1$ to $\epsilon=20$. 
We decide to use the Mean Average Percentage Error (MAPE) as loss function,
\begin{align}
L_{\bm{\mathcal{Y}}_n^{(i)},t}&=\frac{1}{|t|}\sum\limits_{(\bm{U},a_0)\ \in\ t\ \subseteq\ T_{1/2}^{(i-1)}} \left|\frac{\bm{\mathcal{Y}}_n^{(i)}(\bm{U})-a_0}{a_0}\right|, \notag \\ L_{\bm{\mathcal{Y}}^{(i)},t}&=\frac{1}{|t|}\sum\limits_{(\bm{U},a_0)\ \in\ t\ \subseteq\ T_{1/2}^{(i-1)}} \frac{\left|\frac{1}{N_i}\sum\limits_{i=1}^{N_i}\bm{\mathcal{Y}}_n^{(i)}(\bm{U})-a_0\right|}{\left|a_0\right|}. \notag
\end{align}
The upper expression is used to evaluate the loss of a single member $\bm{\mathcal{Y}}_n^{(i)}$ during 
training, while the lower expression corresponds to the MAPE of the entire ensemble.
Processing these losses and computing corresponding weight updates by the Adam optimizer, see 
Refs.~\cite{Kingma:2017kb} and \cite{Loshchilov:2019lh}, we perform mini-batch learning 
with batch size $B=128$.

When it comes to the Taylor decomposition of the ensembles $\bm{\mathcal{Y}}^{(i)}$, it is convenient 
to choose the same expansion point, that is $\bm{U}=\bm{0}$, as we have already seen for $a_0$ in 
Sec.~\ref{sec:varderiv}. We note that the scattering length $a_0(\bm{U})=0$ vanishes in the case with 
no interaction. Therefore, we expect the dominating term of the Taylor series of 
$\bm{\mathcal{Y}}^{(1)}$ not to be the first, constant term, but the second summand, containing the 
first-order derivatives $\partial \bm{\mathcal{Y}}_n^{(1)} / \partial U_k |_{\bm{U}=\bm{0}}$. With 
$(\bm{U}, a_0(\bm{U}))\in T_{1/2}^{(0)}$ this motivates us to skip one iteration and to directly 
perform the substraction
\begin{equation}
a_0^\prime(\bm{U}) = a_0(\bm{U}) - \frac{1}{N_1}\sum\limits_{n=1}^{N_1}\bm{\mathcal{Y}}_n^{(1)}(\bm{0}) - \frac{1}{N_1}\sum\limits_{n=1}^{N_1}\sum\limits_{k=1}^{H_0} \left. \frac{\bm{\mathcal{Y}}_n^{(1)}}{\partial U_k} \right|_{\bm{U}=\bm{0}} U_k \label{eq-newtargets-x3}
\end{equation}
in order to compute samples $(\bm{U}, a_0^\prime(\bm{U}))\in T_{1/2}^{(1)}$ of the successive data 
sets. We expect these new targets to have a vanishing constant and linear contribution and, therefore, 
to behave mainly like $1/2 \cdot \bm{U}^\top K \bm{U}$ with the Hessian $K\in\mathbb{R}^{H_0\times H_0}$. 
Since we are already satisfied with these first two orders, we stop after training the auxiliary 
ensemble $\bm{\mathcal{Y}}^{(2)}$ on $T_1^{(1)}$, using the same training pipeline as before, and do not 
perform further iterations beyond that.

\begin{figure}[t]
\includegraphics[width=0.5\textwidth]{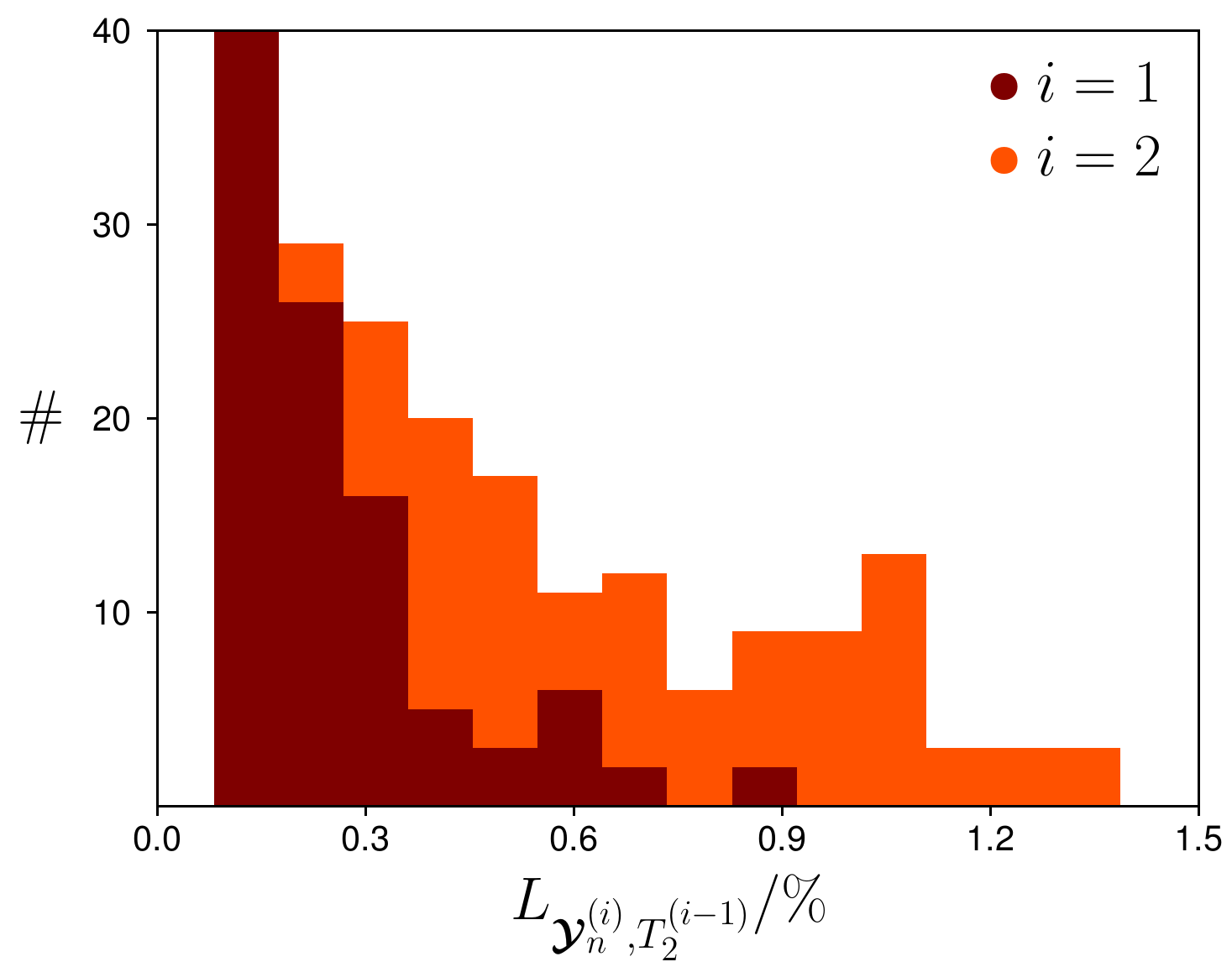}
\caption{Histogram of individual MAPEs $L_{\bm{\mathcal{Y}}_n^{(1)}, T_{2}^{(0)}}$ and $L_{\bm{\mathcal{Y}}_n^{(2)}, T_{2}^{(1)}}$ among all members of the ensembles $\bm{\mathcal{Y}}^{(1)}$ and $\bm{\mathcal{Y}}^{(2)}$ with respect to the corresponding test sets $T_{2}^{(0)}$ and $T_{2}^{(1)}$. We see that members of the first ensemble perform slightly better, which also leads to a better MAPE $L_{\bm{\mathcal{Y}}^{(1)}, T_{2}^{(0)}}$ for the entire ensemble $\bm{\mathcal{Y}}^{(1)}$. Since all hyperparameters are drawn from the same distributions for both ensembles, it appears that it is an easier task to learn a linear than a quadratic relation.}
\label{fig-mape-dist}
\end{figure}

Both ensembles perform sufficiently well on their respective test sets, which follows from their low 
MAPEs of $L_{\bm{\mathcal{Y}}^{(1)},T_2^{(0)}}= 0.152\%$ and 
$L_{\bm{\mathcal{Y}}^{(2)},T_2^{(1)}}=0.438\%$. Note that $\bm{\mathcal{Y}}^{(2)}$ as 
well as its individual members exhibits a slightly worse performance than the members of 
$\bm{\mathcal{Y}}^{(1)}$, cf. Fig.~\ref{fig-mape-dist}. Since both ensembles draw their hyperparameters 
from the same probability distributions, the task of adapting to a dominating, quadratic relation 
between inputs and targets appears to be more challenging that learning a constant or linear relation.

\vfill

\section{First-Order Born Term}
\label{sec:born1}
Given the first ensemble $\bm{\mathcal{Y}}^{(1)}$, we first verify that its members have, indeed, 
adapted to a vanishing axis intercept. This is an important performance requirement, as the scattering 
length vanishes in the force-free case $\bm{U}=0$, which we choose as an expansion point for our proxy 
model. Deriving errors of ensemble-related quantities by computing the standard deviation of that 
quantity among all members, we find
\begin{equation}
\bm{\mathcal{Y}}^{(1)}(\bm{0}) = \frac{1}{N_1}\sum\limits_{n=1}^{N_1}\bm{\mathcal{Y}}_n^{(1)}(\bm{0})=-(1.51\pm 2.38)\times 10^{-4}. \label{eq-axis-intercept-1-x3}
\end{equation}
As $\bm{\mathcal{Y}}^{(1)}(\bm{0})$ takes a small value, compared to the range of all targets in 
$T_{1/2}^{(0)}$, and since the corresponding error even has a slightly larger magnitude, we can confirm 
a vanishing axis intercept for the first ensemble.

\begin{figure}[t]
\includegraphics[width=0.5\textwidth]{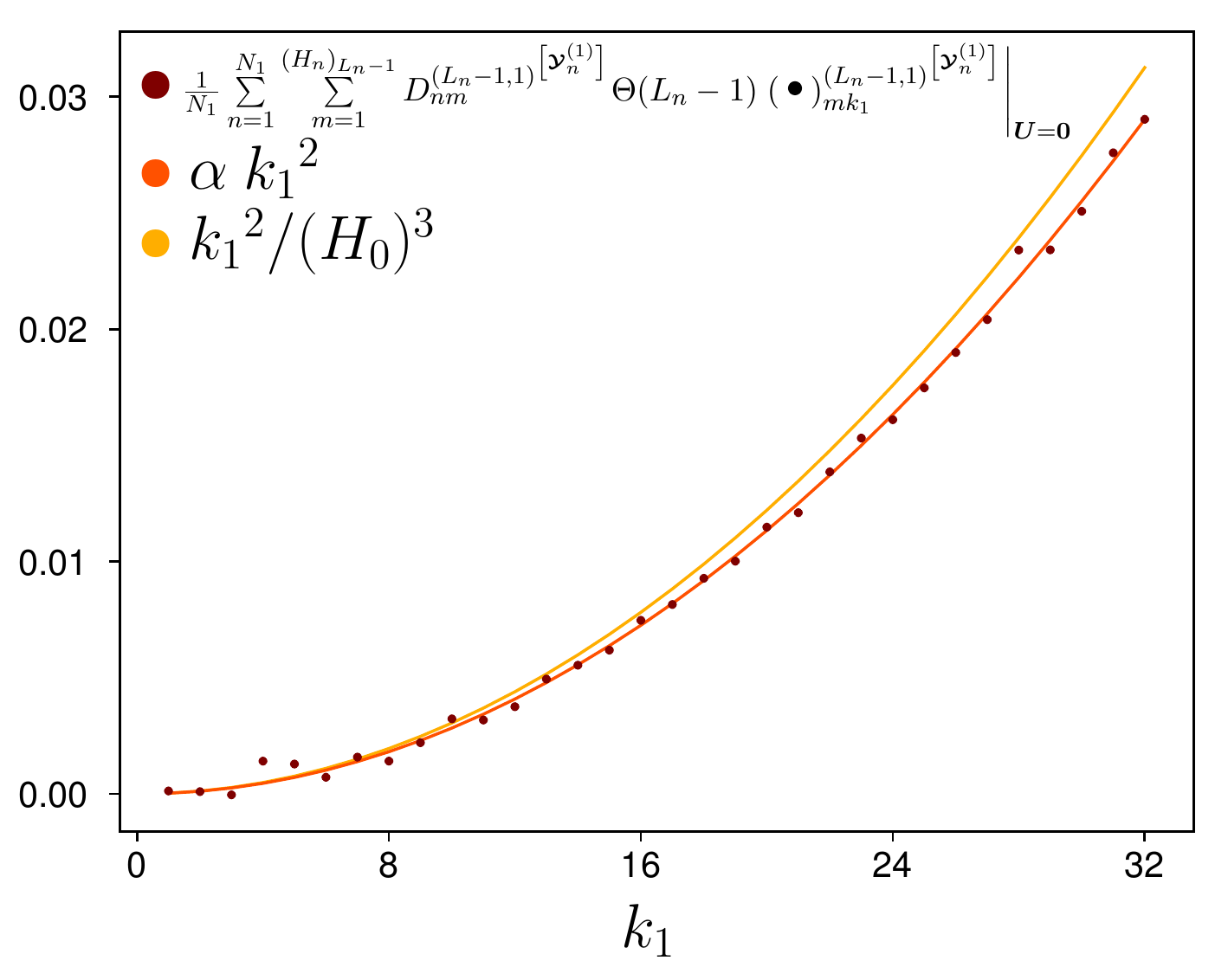}
\caption{First-order Taylor coefficients of the ensemble $\bm{\mathcal{Y}}^{(1)}$, values of the model 
$\alpha\ k_1{}^2$ fitted over these coefficients and first-order Taylor coefficients $k_1{}^2/(H_0)^3$ 
of the sampled Born series over the index $k_1$. As $\alpha$ deviates just slightly more than $1\sigma$ 
from $1/(H_0)^3$, this ensemble, indeed, applies the first-order Born approximation to shallow potentials 
in order to predict S-wave scattering lengths.}
\label{fig-born1-fit-x3}
\end{figure}

The next step is crucial, not only to this iteration but also to the success of the following one: We 
compute the first-order Taylor coefficients of the ensemble $\bm{\mathcal{Y}}^{(1)}$ using 
Eq.~\eqref{eq-taylor1-x2} and the formalism provided in Sec.~\ref{sec:pt}. This reveals its dominating, 
linear contribution in the space of sampled potentials. Ideally, the ensemble would reproduce the linear 
contribution in Eq.~\eqref{eq-born-kernel-x3} of the scattering length one-to-one, which then would 
imply for the Taylor coefficients
\begin{equation}
\left. \frac{\partial \bm{\mathcal{Y}}^{(1)}}{\partial U_{k_1}}\right|_{\bm{U}=\bm{0}}=\frac{1}{N_1}\sum\limits_{n=1}^{N_1}\sum\limits_{m=1}^{(H_n)_{L_n-1}} D_{nm}^{(L_n-1,1)^{\left[\bm{\mathcal{Y}}_n^{(1)}\right]}} \Theta(L_n-1)\left. \left(\ \begin{tikzpicture}[baseline=-2pt, inner sep=0pt, outer sep=4pt]
    \begin{feynman}
        \vertex [dot] (k1) at (0,0) {};
        \diagram*{
            (k1) -- (k1)
        };
    \end{feynman}
\end{tikzpicture}\vphantom{\sum}\ \right)_{mk_1}^{(L_n-1,1)^{\left[\bm{\mathcal{Y}}_n^{(1)}\right]}} \right|_{\bm{U}=\bm{0}} = \frac{k_1{}^2}{(H_0)^3} \label{eq-born1-ideal-x3}
\end{equation}
The superscript $\left[\bm{\mathcal{Y}}_n^{(1)}\right]$ points out that the respective quantity is 
computed for the weights and biases of the ensemble member $\bm{\mathcal{Y}}_n^{(1)}$. In order to 
evaluate how well the left-hand and right-hand sides actually match, we fit the model $\alpha k_1{}^2$ 
to the $H_0=32$ ensemble Taylor coefficients on the left-hand side. Considering the mean and standard 
deviation of the distribution of all fitting parameters among the members of $\bm{\mathcal{Y}}^{(1)}$ 
yields
\begin{equation}
\alpha = (2.834\pm 0.169)\times 10^{-5}.
\end{equation}
The ensemble's Taylor coefficients are displayed together with the fitted curve and the values 
$k_1{}^2/(H_0)^3$ in Fig.~\ref{fig-born1-fit-x3}.
We note that the deviation of the fitting parameter $\alpha$ from the value 
$1/(H_0)^3=3.052\times 10^{-5}$ is just slightly larger than~$1\sigma$. This shows that the ensemble 
$\bm{\mathcal{Y}}^{(1)}$ reproduces the first-order Born approximation sufficiently well and, thereby, 
predicts S-wave scattering lengths for shallow potentials.

\section{Second-Order Born Term}
\label{sec:born2}
Having identified and analyzed the linear contribution of the first ensemble $\bm{\mathcal{Y}}^{(1)}$, 
it is time to move over to the successive data sets $T_{1/2}^{(1)}$ with their targets derived according 
to Eq.~\eqref{eq-newtargets-x3} and to train the auxiliary ensemble $\bm{\mathcal{Y}}^{(2)}$. As argued 
in Sec.~\ref{sec:network}, it is the quadratic contribution, based on the Hessian, which dominates these 
new targets. Similar to the investigation of the first-order Taylor coefficients in Sec.~\ref{sec:born1}, 
we can specify an ideal case in which $\bm{\mathcal{Y}}^{(2)}$ would reproduce the second-order Born term 
one-to-one, namely if their second-order Taylor coefficients satisfy
\begin{equation}\begin{aligned}\left. \frac{\partial^2\bm{\mathcal{Y}}^{(2)}}{\partial U_{k_1}\ \partial U_{k_2}}\right|_{\bm{U}=\bm{0}} &= 
 {\displaystyle \frac{1}{N_2}\sum\limits_{n=1}^{N_2}\sum\limits_{m=1}^{(H_n)_{L_n-1}}}\left[ D_{nm}^{(L_n-1,1)^{[\bm{\mathcal{Y}}_n^{(2)}]}}\Theta(L_n-2)\left( \begin{tikzpicture}[baseline=-2pt, inner sep=0pt, outer sep=4pt]
    \begin{feynman}
        \vertex [dot] (k1) at (0,0) {};
        \vertex [dot] (k2) at (1,0) {};
        \diagram*{
            (k1) -- [dn] (k2)
        };
    \end{feynman}
\end{tikzpicture}\vphantom{\sum}\right)_{mk_1k_2}^{(L-1,2)^{[\bm{\mathcal{Y}}_n^{(2)}]}} \right. \\  &\ \hspace{1cm} + \left. \left. D_{nm}^{(L_n-1,2)^{[\bm{\mathcal{Y}}_n^{(2)}]}}\Theta(L_n-1)\left(\ \begin{tikzpicture}[baseline=-2pt, inner sep=0pt, outer sep=4pt]
    \begin{feynman}
        \vertex [dot] (k1) at (0,0) {};
        \diagram*{
            (k1) -- (k1)
        };
    \end{feynman}
\end{tikzpicture}\vphantom{\sum}\ \right)_{mk_1}^{(L-1,1)^{[\bm{\mathcal{Y}}_n^{(2)}]}}\left(\ \begin{tikzpicture}[baseline=-2pt, inner sep=0pt, outer sep=4pt]
    \begin{feynman}
        \vertex [dot] (k1) at (0,0) {};
        \diagram*{
            (k1) -- (k1)
        };
    \end{feynman}
\end{tikzpicture}\vphantom{\sum}\ \right)_{mk_2}^{(L-1,1)^{[\bm{\mathcal{Y}}_n^{(2)}]}} \right] \right|_{\bm{U}=\bm{0}} \\
&= -\frac{1}{(H_0)^5} k_1 k_2 \left(k_1 + k_2 - |k_1-k_2|\right).
\end{aligned}\label{eq-born2-ana}
\end{equation}
In order to investigate, how close the auxiliary ensemble actually approximates this ideal case, we fit 
the model $\beta\ k_1 k_2 \left(k_1 + k_2 - |k_1-k_2|\right)$. If we observe the fitting parameter $\beta$ 
to closely approach the value $-1/(H_0)^5=-2.98\times 10^{-8}$, we can be certain that 
$\bm{\mathcal{Y}}^{(2)}$ applies the second-order Born term to predict the dominating, quadratic 
contribution. 

\begin{figure}[t]
\includegraphics[width=\textwidth]{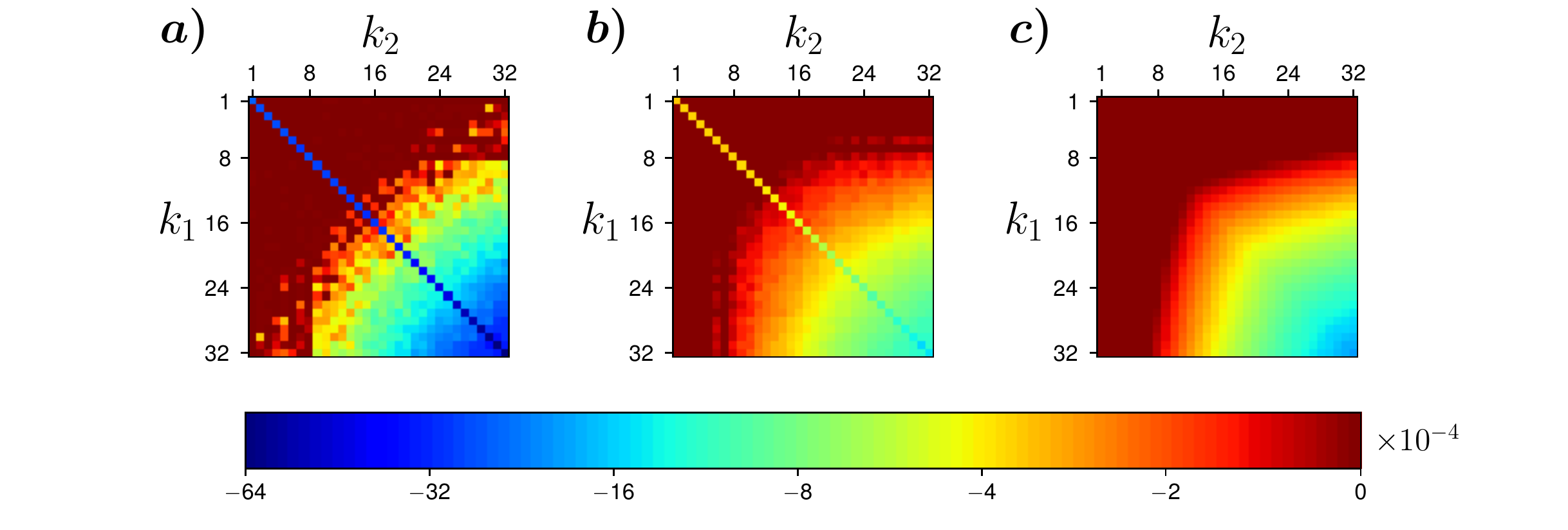}
\caption{Second-order Taylor coefficients of the ensembles $\bm{\mathcal{Y}}^{(1)}$ (Fig.~$a$), $\bm{\mathcal{Y}}^{(2)}$ (Fig.~$b$) and the Hessian of the Born series (Fig.~$c$). Both ensembles reproduce the basic behavior displayed of the Hessian in Fig.~$c)$. Since adapting to the second-order Born term has not been prioritized during the training of $\bm{\mathcal{Y}}^{(1)}$, the resulting elements in Fig.~$a)$ are noisy and very large values appear in the bottom right corner. In contrast to $\bm{\mathcal{Y}}^{(1)}$, the auxiliary ensemble $\bm{\mathcal{Y}}^{(2)}$ has adapted to the second-order Born term much better. The diagonals appearing on both ensemble Hessians is presumably an artifact that could be eliminated using other and more capable architectures than MLPs.}
\label{fig-born2-x3}
\end{figure}

To justify our perturbative ansatz, we not only compute the Hessian of the auxiliary ensemble 
$\bm{\mathcal{Y}}^{(2)}$, but also consider the Hessian of the ensemble $\bm{\mathcal{Y}}^{(1)}$ from 
the previous iteration. The latter can be expected to be significantly less faithful to the second-order 
Born term, since the quadratic contribution to scattering lenghs of shallow potentials is lower than 
that of the linear term from Sec.~\ref{sec:born1} and, thus, has not been prioritized during training. 
Both Hessians and the Hessian of the Born series, c.f. Eq.\eqref{eq-born-kernel-x3}, are shown in 
Fig.~\ref{fig-born2-x3}. For $\bm{\mathcal{Y}}^{(2)}$ we find the fitting parameter 
$\beta = (-2.25 \pm 1.63)\times 10^{-8}$. Indeed, the deviation from $-1/(H_0)^5$ is less than 
$1\sigma$. But at this point, we also notice the unfortunately large error, which may be explained by 
the slightly weaker performance of the auxiliary ensemble. Moreover, interestingly, we can observe a 
very distinct diagonal in both ensemble Hessians, which does not appear in the second-order Born term. 
Since weight decay and ensembling have a regulatory influence on the resulting predictions, we can 
exclude overfitting as cause. We, therefore, conjecture that these diagonals are artifacts that might 
disappear when using other, more capable architectures than MLPs. Apart from that, 
Fig.~\ref{fig-born2-x3}$b)$ shows that the auxiliary ensemble $\bm{\mathcal{Y}}^{(2)}$ mostly reproduces 
the desired behavior. And a glimpse on Fig.~\ref{fig-born2-x3}$a)$ lets us surmise that even the ensemble 
$\bm{\mathcal{Y}}^{(1)}$ very roughly behaves like $-1/(H_0)^5k_1k_2\left(k_1 + k_2 - |k_1-k_2|\right)$. 
Nonetheless, by moving to the auxiliary ensemble, much of the noise attached to the coefficients in 
Fig.~\ref{fig-born2-x3}$a)$ is heavily reduced and the desired shape of the Hessian displayed in 
Fig.~\ref{fig-born2-x3}$c)$ is much better approximated. In conclusion, we consider the Hessian of the 
auxiliary ensemble to be suitable for constructing an S-wave scattering length proxy.

\begin{figure}[t]
\includegraphics[width=0.5\textwidth]{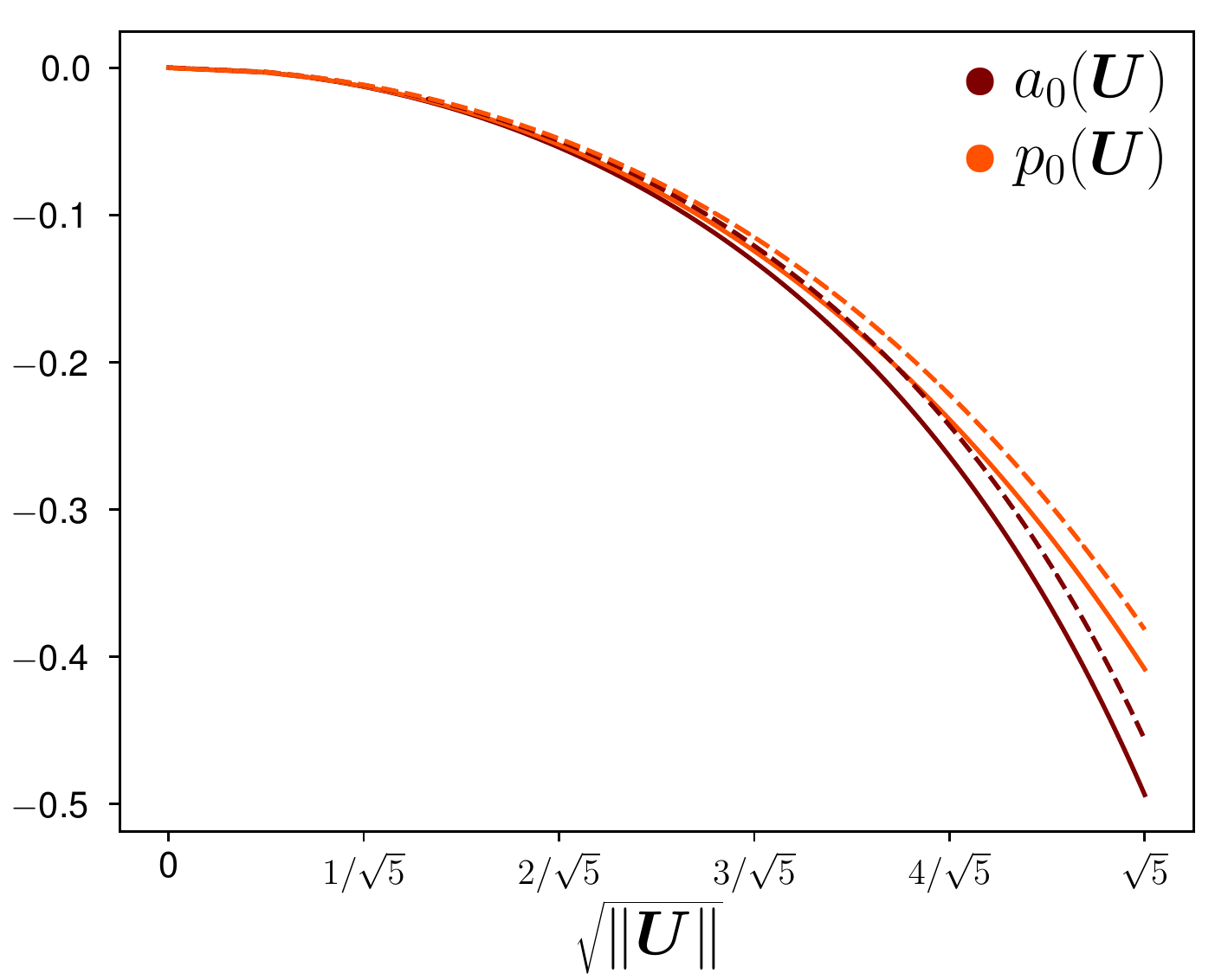}
\caption{Scattering lengths $a_0(\bm{U})$ and corresponding scattering length predictions $p_0(\bm{U})$ by the proxy from Eq.~\eqref{eq-proxyfinal-x3} for 200 different, sampled potentials that take one of two randomly generated shapes. Solid (dashed) lines are related to potentials of the first (second) randomly generated shape. The relative error between $p_0(\bm{U})$ and $a_0(\bm{U})$ is less than $3\%$ for shallow potentials with $\|\bm{U}\| \lesssim 1$. Introducing successively higher-order terms to the proxy would provide better predictions and would, consequently, increase the range of validity.}
\label{fig-proxy}
\end{figure}

In completing the second iteration, we can now use the gradient and the vanishing axis intercept of the 
first ensemble $\bm{\mathcal{Y}}^{(1)}$ and the Hessian of the auxiliary ensemble $\bm{\mathcal{Y}}^{(2)}$ 
to construct a much simpler proxy
\begin{equation}
p_0(\bm{U}) = \bm{\mathcal{Y}}^{(1)}(\bm{0}) + \sum\limits_{k_1=1}^{H_0} \left. \frac{\partial \bm{\mathcal{Y}}^{(1)}}{\partial U_{k_1}} \right|_{\bm{U}=\bm{0}}U_{k_1} + \frac{1}{2} \sum\limits_{k_1=1}^{H_0}\sum\limits_{k_2=1}^{H_0} \left. \frac{\partial^2 \bm{\mathcal{Y}}^{(2)}}{\partial U_{k_1} \partial U_{k_2}} \right|_{\bm{U}=\bm{0}}U_{k_1}U_{k_2}. \label{eq-proxyfinal-x3}
\end{equation}
The proxy $p_0$ can be understood as a machine-learned second-order Born approximation. In order to examine 
its range of validity, we proceed as follows: At first we randomly generate two different
potential shapes $\bm{n}_i\in\mathbb{R}^{H_0}$. For both of these shapes we generate a set of $100$
equidistant potentials $\bm{U}_i=\|\bm{U}_i\|\bm{n}_i$ with magnitudes $\|\bm{U}_i\|\in[0,\ldots,5]$.
For each of these potentials $\bm{U}$ the above, machine-learned Born approximation $p_0(\bm{U})$ and true 
scattering lengths $a_0(\bm{U})$ are evaluated and
plotted in Fig.~\ref{fig-proxy}. We observe that the relative error between $p_0(\bm{U})$ and $a_0(\bm{U})$ is less 
than $3\%$ for shallow potentials with $\|\bm{U}\| \lesssim 1$. Beyond that regime, additional higher-order 
terms must be introduced to the proxy to make better scattering length predictions.

\section{Discussion and Outlook}
\label{sec:summ}
In this manuscript we propose a neural network perturbation theory for MLPs. This allows us to construct 
much simpler proxies of the original target function. The key idea of that perturbation theory is the 
successive identification and elimination of the leading-order contribution to the ensemble's Taylor 
decomposition. This establishes a sequence of ensembles that each specialize in approximating consecutive 
orders of the target function's Taylor decomposition. Combining these accordingly can, thus, be viewed as 
the first step of a proxy method, see Ref.~\cite{Fan:2020fxw}. 

Especially when dealing with deep MLPs, the computation of higher-order Taylor coefficients can be a 
challenging task. Nonetheless, we manage to obtain an analytical expression for partial derivatives of any 
order for arbitrarily deep, analytically activated MLPs in terms of propagators and vertices. The underlying 
formalism is motivated by Feynman diagrams in quantum field theories and the entailed graph theoretical 
approach makes the underlying combinatorics significantly more systematical and manageable. Note that the 
graphical representation of its derivatives does not depend on the particular choice of an MLP:
Indeed, the calculation of propagators, vertices and saturation thresholds themselves may be altered due 
to varying 
weights, biases, activation functions, number of neurons and hidden layers. However,
there is no way to uniquely infer more information about its architecture from the mere structure of the 
contributing arborescences.

We apply this graphical formalism and neural network perturbation theory to S-wave scattering lengths of 
shallow potentials. For this, we train two ensembles within the same number of iterations. Using the axis 
intercept and gradient of the first ensemble and the Hessian from the second, auxiliary ensemble yields a 
proxy of the Born series, which reproduces the second-order Born approximation for shallow potentials. At 
this point one could, of course, argue that it would have been much more convenient to simply
train an NN $\bm{\mathcal{Y}}:\mathbb{R}^{32}\to\mathbb{R}$ that is just the sum of one linear layer $\bm{l}$
and one bilinear layer $B$, that is
\begin{equation}
\bm{\mathcal{Y}}(\bm{U}) = \bm{l\cdot U} + \frac{1}{2} \bm{U}^\top B \bm{U}.\notag
\end{equation}
Using this architecture instead of deep MLPs would not only reduce the computational effort
significantly, but also would have imposed some desired properties like $\bm{\mathcal{Y}}(\bm{0})=0$ and
simultaneously learning the first- and second-order Born terms. Note that $\bm{\mathcal{Y}}$ in this case is a
second-order Taylor approximation by itself, which allows to directly read off Taylor coefficients
instead of deriving them first, as performed in our analysis. However, such an NN is not an
universal approximator, as it violates the UAT, and, therefore,
will fail in reproducing scattering lengths for deeper potentials than in this analysis.
This is because models of this architecture are unable to adapt to higher-order terms of
the Born series. Therefore, using such an architecture may indeed simplify the
analysis, but must be well justified for the particular case.

Note that the obvious next step of this analysis would be the interpretation of the constructed
proxy in the space of all sampled potentials. In doing so, we would just gather a post-hoc
interpretation, based on approximations and thus deviations
from actual scattering lengths.
In this case, prediction and interpretation, therefore, have
to be understood as two independent instances. In recent years there have been many efforts
to close the gap between prediction and interpretation by ad-hoc interpretation methods.
These exemplarily involve training NNs whose architectures are either intrinsically
interpretable or can be brought in an interpretable representation, see Ref.~\cite{Fan:2020fxw}.
At the cost of a prediction-interpretation-tradeoff, the advantage of ad-hoc methods is
that resulting intepretations are completely faithful to the NN's prediction, in
contrast to the mentioned post-hoc methods.

\section*{Acknowledgements}
We are grateful to J\"urgen Gall  as well as the reviewers for useful comments.
  We acknowledge partial financial support by the Deutsche Forschungsgemeinschaft (DFG, German Research
Foundation) and the NSFC through the funds provided to the Sino-German Collaborative  
Research Center TRR110 ``Symmetries and the Emergence of Structure in QCD''
(DFG Project-ID 196253076 - TRR 110, NSFC Grant No. 12070131001).
Further support was provided by the Chinese Academy of Sciences (CAS) President's International
Fellowship Initiative (PIFI) (grant no. 2018DM0034), by VolkswagenStiftung (grant no. 93562) and
by the EU (Strong2020).\\

\clearpage

\section*{Appendix A\ \ \ $-$\ \ \ Proofs}
\subsection{Proof of Eq.~\eqref{eq-propderiv-1}}
\label{sec-easyproof}
\textbf{Theorem 1.} \emph{The first-order partial derivatives of the $nm^\text{th}$ matrix element of the $l^\text{th}$ layer propagator $D_{nm}^{(l,p)}$ of order $p$ is given by}
\vfill
\noindent
\begin{equation}
\frac{\partial D_{nm}^{(l,p)}}{\partial x_k} = D_{nm}^{(l,p+1)} \Delta_{mk}^{(l,1)}, \notag
\end{equation} 
\vfill
\noindent
where we have introduced the matrix elements
\vfill
\noindent
\begin{equation}
\Delta_{mk}^{(l,1)}=\sum\limits_{q_l=1}^{H_l}\ldots\sum\limits_{q_1=1}^{H_1}\delta_{mq_l} w_{q_1 k}^{(1)}\prod\limits_{i=1}^{l-1}D_{q_{i+1}q_{i}}^{(i,1)}. \notag
\end{equation}
\vfill
\noindent
\emph{Proof.} First of all, it is easy to see that the derivative with respect to the $k^\text{th}$ component $x_k$ of the input is proportional to a propagator of higher order $p+1$. Due to the chain rule, the term $\partial z_m^{(l)}/\partial x_k$ appears, 
\vfill
\noindent
\begin{align}
\frac{\partial D_{nm}^{(l,p)}}{\partial x_k} &= \underbrace{w_{nm}^{(l+1)}\frac{\mathrm{d}^{p+1}a^{(l,m)}}{\mathrm{d}x^{p+1}}\left( z^{(l)}_m \right)}_{=D_{nm}^{(l,p+1)}}\sum\limits_{q_l=1}^{H_l}\delta_{mq_l}\frac{\partial z_{q_l}^{(l)}}{\partial x_k}. \notag
\end{align}
\vfill
\noindent
By inserting the recursive step from Eq.~\eqref{eq-mlprecursion}, this dependency can be shifted to the previous layer,
\vfill
\noindent
\begin{align}
\frac{\partial D_{nm}^{(l,p)}}{\partial x_k}&= D_{nm}^{(l,p+1)} \sum\limits_{q_l=1}^{H_l}\delta_{mq_l}\sum\limits_{q_{l-1}=1}^{H_{l-1}} w_{q_{l}q_{l-1}}^{(l)}\frac{\partial y_{q_{l-1}}^{(l-1)}}{\partial x_k} \notag \\
&= D_{nm}^{(l,p+1)} \sum\limits_{q_l=1}^{H_l}\delta_{mq_l}\sum\limits_{q_{l-1}=1}^{H_{l-1}} \underbrace{w_{q_{l}q_{l-1}}^{(l)}\frac{\mathrm{d}a^{(l-1,q_{l-1})}}{\mathrm{d}x}\left( z^{(l-1)}_{q_{l-1}} \right)}_{=D_{q_{l}q_{l-1}}^{(l-1,1)}}\frac{\partial z_{q_{l-1}}^{(l-1)}}{\partial x_k}. \notag 
\end{align}
\vfill
\noindent
In the same manner, we can apply the chain rule successively to all antecedent layers until the base $y_{q_0}^{(0)}~=~x_{q_0}$ is reached. Thereby, each layer provides a matrix-multiplication with a first-order propagator:
\vfill
\noindent
\begin{align}
\frac{\partial D_{nm}^{(l,p)}}{\partial x_k} &= D_{nm}^{(l,p+1)} \sum\limits_{q_l=1}^{H_l}\delta_{mq_l}\sum\limits_{q_{l-1}=1}^{H_{l-1}} D_{q_{l}q_{l-1}}^{(l-1,1)}\sum\limits_{q_{l-2}=1}^{H_{l-2}} D_{q_{l-1}q_{l-2}}^{(l-2,1)}\ldots \sum\limits_{q_{1}=1}^{H_{1}} D_{q_{2}q_{1}}^{(1,1)}\sum\limits_{q_0=1}^d w_{q_1q_0}^{(1)}\underbrace{\frac{\partial x_{q_0}}{\partial x_k}}_{=\delta_{q_0k}}. \notag
\end{align}
\vfill
\noindent
Rearranging those propagators and sums finally yields
\vfill
\noindent
\begin{align}
\frac{\partial D_{nm}^{(l,p)}}{\partial x_k} &= D_{nm}^{(l,p+1)}\sum\limits_{q_l=1}^{H_l}\ldots\sum\limits_{q_1=1}^{H_1}\delta_{mq_l} w_{q_1 k}^{(1)}\prod\limits_{i=1}^{l-1}D_{q_{i+1}q_{i}}^{(i,1)}. \notag
\end{align}
\clearpage
\subsection{Proof of Eq.~\eqref{eq-propderiv-2}}
\label{sec-proofs}
\noindent
\textbf{Theorem 2 (Eq.~\eqref{eq-propderiv-2}).} \emph{The $N^\text{th}$ derivative of the propagator $D_{nm}^{(l,p)}$ is given by}
\vfill
\noindent
\begin{equation}
\frac{\ \hspace{0.1cm}\partial^N D_{nm}^{(l,p)}}{\partial x_{k_1}\ldots\partial x_{k_N}} = \sum\limits_{c=1}^ND_{nm}^{(l,p+c)}\sum\limits_{\sigma\in\mathcal{S}_N}\sum\limits_{\bm{\pi}\in \Pi^c_N}\frac{1}{\varepsilon_{\bm{\pi}}}\prod\limits_{i=1}^c \Delta_{mk_{\sigma\left(1+\sum_{j=1}^{i-1}\pi_j\right)}\ldots\ k_{\sigma\left(\sum_{j=1}^{i}\pi_j\right)}}^{(l,\pi_i)}.\notag
\end{equation}
\vfill
\noindent
\emph{Proof.} We prove Eq.~\eqref{eq-propderiv-2} by a complete induction. Therefore, we quickly convince ourselves of its validity in the base case $N=1$ with $\varepsilon_{(1)}=1$,
\vfill
\noindent
\begin{align}
\frac{\ \hspace{0.1cm}\partial D_{nm}^{(l,p)}}{\partial x_{k_1}} &= D_{nm}^{(l,p+1)}\hspace{-0.2cm}\sum\limits_{(\pi_1)\ \in\ \{(1)\}}\frac{1}{\varepsilon_{(1)}}\sum\limits_{\sigma\in\{\mathrm{id}\}}\Delta_{mk_{\sigma\left(\pi_1\right)}}^{(l,\pi_1)}\notag\\
&= D_{nm}^{(l,p+1)} \Delta_{mk_1}^{(l,1)}.\notag
\end{align}
\vfill
\noindent
This, indeed, corresponds to Eq.~\eqref{eq-propderiv-1}, which is also the defining equation for $\Delta_{mk_1}^{(l,1)}$. Subsequently, the inductive step involves evaluating the derivative
\vfill
\noindent
\begin{equation}
\frac{\ \hspace{0.1cm}\partial^{N+1} D_{nm}^{(l,p)}}{\partial x_{k_1}\ldots\partial x_{k_{N+1}}} = \frac{\partial}{\partial x_{k_{N+1}}}\sum\limits_{c=1}^{N}D_{nm}^{(l,p+c)}\sum\limits_{\sigma\in\mathcal{S}_N}\sum\limits_{\bm{\pi}\in \Pi^c_N}\frac{1}{\varepsilon_{\bm{\pi}}}\prod\limits_{i=1}^c \Delta_{mk_{\sigma\left(1+\sum_{j=1}^{i-1}\pi_j\right)}\ldots\ k_{\sigma\left(\sum_{j=1}^{i}\pi_j\right)}}^{(l,\pi_i)}.\notag
\end{equation}
\vfill
\noindent
By applying the product rule, we either encounter propagator derivatives or derivatives of tensor elements, which we split into two distinct sums,
\vfill
\noindent
\begin{align}
\frac{\ \hspace{0.1cm}\partial^{N+1} D_{nm}^{(l,p)}}{\partial x_{k_1}\ldots\partial x_{k_{N+1}}} &\ \hspace{-0.35cm}\overset{\text{\eqref{eq-propderiv-1}, \eqref{eq-delta-1}}}{=}\sum\limits_{c=1}^ND_{nm}^{(l,p+c+1)}\Delta_{mk_{N+1}}^{(l,1)}\hspace{-0.2cm}\sum\limits_{\sigma\in\mathcal{S}_N}\sum\limits_{\bm{\pi}\in \Pi^c_N}\frac{1}{\varepsilon_{\bm{\pi}}}\prod\limits_{i=1}^c \Delta_{mk_{\sigma\left(1+\sum_{j=1}^{i-1}\pi_j\right)}\ldots\ k_{\sigma\left(\sum_{j=1}^{i}\pi_j\right)}}^{(l,\pi_i)}\notag\\
&\ \ \ +\sum\limits_{c=1}^ND_{nm}^{(l,p+c)}\hspace{-0.2cm}\sum\limits_{\sigma\in\mathcal{S}_N}\sum\limits_{\bm{\pi}\in \Pi^c_N}\frac{1}{\varepsilon_{\bm{\pi}}}\sum\limits_{i=1}^c \Delta_{mk_{\sigma\left(1+\sum_{j=1}^{i-1}\pi_j\right)}\ldots\ k_{\sigma\left(\sum_{j=1}^{i}\pi_j\right)}k_{N+1}}^{(l,\pi_i+1)}\notag\\
&\ \hspace{4.7cm} \times \prod\limits_{\mbox{\scriptsize $ \begin{matrix}i^\prime=1\\ i^\prime\neq i \end{matrix} $ }}^c \Delta_{mk_{\sigma\left(1+\sum_{j=1}^{i^\prime-1}\pi_j\right)}\ldots\ k_{\sigma\left(\sum_{j=1}^{i^\prime}\pi_j\right)}}^{(l,\pi_{i^\prime})}.\notag
\end{align}
\vfill
\noindent
Up to the $N^\text{th}$ summand in the first sum and the first summand in the second sum, all other summands can be combined to one sum from $c=2$ to $c=N$,
\vfill
\noindent
\begin{align}
\frac{\ \hspace{0.1cm}\partial^{N+1} D_{nm}^{(l,p)}}{\partial x_{k_1}\ldots\partial x_{k_{N+1}}}  &=D_{nm}^{(l,p+N+1)}\Delta_{mk_{N+1}}^{(l,1)}\hspace{-0.2cm}\sum\limits_{\sigma\in\mathcal{S}_N}\sum\limits_{\bm{\pi}\in \Pi^N_N}\frac{1}{\varepsilon_{\bm{\pi}}}\prod\limits_{i=1}^N \Delta_{mk_{\sigma\left(1+\sum_{j=1}^{i-1}\pi_j\right)}\ldots\ k_{\sigma\left(\sum_{j=1}^{i}\pi_j\right)}}^{(l,\pi_i)}\notag
\\
&\ \ +\sum\limits_{c=2}^ND_{nm}^{(l,p+c)}\sum\limits_{\sigma\in\mathcal{S}_N}\left[ \Delta_{mk_{N+1}}^{(l,1)}\hspace{-0.2cm}\sum\limits_{\bm{\alpha}\in \Pi^{c-1}_N}\frac{1}{\varepsilon_{\bm{\alpha}}}\prod\limits_{i=1}^{c-1} \Delta_{mk_{\sigma\left(1+\sum_{j=1}^{i-1}\alpha_j\right)}\ldots\ k_{\sigma\left(\sum_{j=1}^{i}\alpha_j\right)}}^{(l,\alpha_i)} \right. \notag
\\
&\ \hspace{4.7cm} + \sum\limits_{\bm{\beta}\in \Pi^{c}_N}\frac{1}{\varepsilon_{\bm{\beta}}}\sum\limits_{i=1}^{c} \Delta_{mk_{\sigma\left(1+\sum_{j=1}^{i-1}\beta_j\right)}\ldots\ k_{\sigma\left(\sum_{j=1}^{i}\beta_j\right)}k_{N+1}}^{(l,\beta_{i}+1)} \notag 
\\
&\ \hspace{6.3cm} \ \ \times\prod\limits_{\mbox{\scriptsize $ \begin{matrix}i^\prime=1\\ i^\prime\neq i \end{matrix} $ }}^{c}\left. \Delta_{mk_{\sigma\left(1+\sum_{j=1}^{i^\prime-1}\beta_j\right)}\ldots\ k_{\sigma\left(\sum_{j=1}^{i^\prime}\beta_j\right)}}^{(l,\beta_{i^\prime})}\right]\notag
\\
&\ \ +D_{nm}^{(l,p+1)}\sum\limits_{\sigma\in\mathcal{S}_N}\sum\limits_{(\pi_1)\in \Pi^{1}_N}\frac{1}{\varepsilon_{(\pi_1)}}\Delta_{mk_{\sigma\left(1\right)}\ldots\ k_{\sigma\left(\pi_1\right)}k_{N+1}}^{(l,\pi_{1}+1)} \notag
\end{align}
\vfill
\noindent
The two external summands can be explicitly derived using $\Pi_N^N=\{(1)_{i=1}^N\}$ and $\Pi_N^1=\{(N)\}$. In both cases, the symmetry factor is given by
\begin{equation}
\varepsilon_{(1,\ldots,1)} = \varepsilon_{(N)} = N!\ \ , \notag
\end{equation}
and, respectively,
\begin{equation}
\varepsilon_{(1,\ldots,1,1)} = \varepsilon_{(N+1)} = (N+1)!\ \ , \notag
\end{equation}
\vfill
\noindent
such that each summand finally can be written as a sum over $\mathcal{S}_{N+1}$:
\begin{align}
\Delta_{mk_{1}}^{(l,1)}\ldots\Delta_{mk_{N+1}}^{(l,1)}&=\Delta_{mk_{N+1}}^{(l,1)}\sum\limits_{\sigma\in\mathcal{S}_N}\sum\limits_{\bm{\pi}\in \Pi^N_N}\frac{1}{\varepsilon_{\bm{\pi}}}\prod\limits_{i=1}^N \Delta_{mk_{\sigma\left(1+\sum_{j=1}^{i-1}\pi_j\right)}\ldots\ k_{\sigma\left(\sum_{j=1}^{i}\pi_j\right)}}^{(l,\pi_i)}\notag\\
&= \sum\limits_{\mbox{\scriptsize $ \begin{matrix}\sigma\in \mathcal{S}_{N+1}\\ \sigma\left(N+1\right)=N+1
\end{matrix} $ }}\sum\limits_{\bm{\pi}\in \Pi^{N+1}_{N+1}}\frac{N+1}{\varepsilon_{\bm{\pi}}}\prod\limits_{i=1}^N \Delta_{mk_{\sigma\left(1+\sum_{j=1}^{i-1}\pi_j\right)}\ldots\ k_{\sigma\left(\sum_{j=1}^{i}\pi_j\right)}}^{(l,\pi_i)}\notag\\
&= \sum\limits_{\sigma\in\mathcal{S}_{N+1}}\sum\limits_{\bm{\pi}\in \Pi^{N+1}_{N+1}}\frac{1}{\varepsilon_{\bm{\pi}}}\prod\limits_{i=1}^N \Delta_{mk_{\sigma\left(1+\sum_{j=1}^{i-1}\pi_j\right)}\ldots\ k_{\sigma\left(\sum_{j=1}^{i}\pi_j\right)}}^{(l,\pi_i)} \notag
\end{align}
\vfill
\noindent
and
\vfill
\noindent
\begin{align}
\Delta_{ml_1\ldots k_{N+1}}^{(l,N+1)} &= \sum\limits_{\sigma\in\mathcal{S}_N}\sum\limits_{(\pi_1)\in\ \Pi^{1}_N}\frac{1}{\varepsilon_{(\pi_1)}}\Delta_{mk_{\sigma\left(1\right)}\ldots\ k_{\sigma\left(\pi_1\right)}k_{N+1}}^{(l,\pi_{1}+1)}\notag\\
&= \sum\limits_{\mbox{\scriptsize $ \begin{matrix}\sigma\in \mathcal{S}_{N+1}\\ \sigma\left(N+1\right)=N+1
\end{matrix} $ }}\sum\limits_{(\pi_1)\in\ \Pi^{1}_{N+1}}\frac{N+1}{\varepsilon_{(\pi_1)}}\Delta_{mk_{\sigma\left(1\right)}\ldots\ k_{\sigma\left(\pi_1\right)}}^{(l,\pi_{1}+1)}\notag\\
&= \sum\limits_{\sigma\in\mathcal{S}_{N+1}}\sum\limits_{(\pi_1)\in\ \Pi^{1}_{N+1}}\frac{1}{\varepsilon_{(\pi_1)}}\Delta_{mk_{\sigma\left(1\right)}\ldots\ k_{\sigma\left(\pi_1\right)}}^{(l,\pi_{1}+1)}
\end{align}
Finally, the remaining sum can be expressed as a sum over $\Pi_{N+1}^c$. Then, combining all terms finally proves the inductive step,
\vfill
\noindent
\begin{align}
\frac{\ \hspace{0.1cm}\partial^{N+1} D_{nm}^{(l,p)}}{\partial x_{k_1}\ldots\partial x_{k_{N+1}}} &= D_{nm}^{(l,p+(N+1))}\sum\limits_{\sigma\in\mathcal{S}_{N+1}}\sum\limits_{\bm{\pi}\in \Pi^{N+1}_{N+1}}\frac{1}{\varepsilon_{\bm{\pi}}}\prod\limits_{i=1}^N \Delta_{mk_{\sigma\left(1+\sum_{j=1}^{i-1}\pi_j\right)}\ldots\ k_{\sigma\left(\sum_{j=1}^{i}\pi_j\right)}}^{(l,\pi_i)} \notag \\
&\ \ \ + \sum\limits_{c=2}^N D_{nm}^{(l,p+c)} \sum\limits_{\sigma\in\mathcal{S}_{N+1}}\sum\limits_{\bm{\pi}\in \Pi^c_{N+1}} \frac{1}{\varepsilon_{\bm{\pi}}} \prod\limits_{i=1}^c \Delta_{mk_{\sigma\left(1+\sum_{j=1}^{i-1}\pi_j\right)}\ldots\ k_{\sigma\left(\sum_{j=1}^{i}\pi_j\right)}}^{(l,\pi_i)}\notag\\
&\ \ \ + D_{nm}^{(l,p+1)}\sum\limits_{\sigma\in\mathcal{S}_{N+1}}\sum\limits_{\bm{\pi}\in \Pi^{1}_{N+1}}\frac{1}{\varepsilon_{\bm{\pi}}}\Delta_{mk_{\sigma\left(1\right)}\ldots\ k_{\sigma\left(\pi_1\right)}}^{(l,\pi_{1}+1)} \notag\\
&=\sum\limits_{c=1}^{N+1} D_{nm}^{(l,p+c)} \sum\limits_{\sigma\in\mathcal{S}_{N+1}}\sum\limits_{\bm{\pi}\in \Pi^c_{N+1}} \frac{1}{\varepsilon_{\bm{\pi}}} \prod\limits_{i=1}^c \Delta_{mk_{\sigma\left(1+\sum_{j=1}^{i-1}\pi_j\right)}\ldots\ k_{\sigma\left(\sum_{j=1}^{i}\pi_j\right)}}^{(l,\pi_i)}\notag.\notag
\end{align}
\vfill
\noindent
\vfill
\subsection{Proof of Eq.~\eqref{eq-delta-sum}}
\label{sec-proofs-2}
\textbf{Theorem 3 (Eq.~\eqref{eq-delta-sum}).} The tensor elements $\Delta_{mk_1\ldots k_N}^{(l,p)}$ can be expressed as the following weighted sum of all $N$-vertex arborescences, as defined in Eq.~\eqref{eq-smalldelta-def}, with adjacency matrices $A\in\mathbb{A}^N$,
\vfill
\noindent
\begin{equation}
\Delta_{mk_1\ldots k_N}^{(l,N)} = \sum\limits_{A\in \mathbb{A}^N} \Theta(l-\alpha(A))\ \delta_{mk_1\ldots k_N}^{(l,N)}(A).\notag
\end{equation}
\vfill
\noindent
Weighting with factors $\Theta(l-\alpha(A))$ causes an arborescence only to contribute, as long as the layer $l$, the tensor element is considered for, overshoots the saturation threshold $\alpha(A)$, given in Eq.~\eqref{eq-alpha-def}.\\\\
\emph{Proof.} Since the base case $(N=1)$ has already been shown in Eq.~\eqref{eq-vertex-1}, we directly start with the inductive step. The commutator formula
\vfill
\noindent
\begin{equation}
\left[ O, \prod\limits_{c=1}^N B_c\right] = \sum\limits_{i=0}^{N-1} \left( \prod\limits_{j=1}^i B_j \right)[O,B_{i+1}]\left( \prod\limits_{j=i+2}^N B_j \right)\label{eq-commutator-1}
\end{equation}
\vfill
\noindent
will later prove to be useful. For $O=\partial/\partial x_{k_{N+1}}$ and for
\vfill
\noindent
\begin{equation}
B_c(A) = \mlpsum\limits_{(q_a^{\{c\}})_{a=1}^{j_c(A)} \vphantom{\begin{matrix}{}\\{}\end{matrix}}\ (j_b^{\{c\}})_{b=1}^{\mathrm{max}_{r}(A_{cr})}}\hspace{-1.5cm}{\vphantom{\mlpsum}}_{q_c(A)\ k_c}\prod\limits_{b=1}^{\mathrm{max}_{r}(A_{cr})} D_{bc}^{(1+n_{bc}(A))}(A)\notag
\end{equation}
\vfill
\noindent
from the $c^\text{th}$ vertex of the arborescence $\delta_{mk_1\ldots k_N}^{(l,N)}(A)$, we derive the individual commutators
\clearpage
\begin{align}
\left[ \frac{\partial}{\partial x_{k_{N+1}}}, B_c(A)\right] &\overset{\text{\eqref{eq-vertexderiv-1}}}{=} \Theta\left(l-\mathrm{max}_{r}(A_{cr})-1\right)\notag
\\
&\ \hspace{2cm}\times\mlpsum\limits_{(q_a^{\{c\}})_{a=1}^{j_c(A)} \vphantom{\begin{matrix}{}\\{}\end{matrix}}\ (j_b^{\{c\}})_{b=1}^{\mathrm{max}_{r}(A_{cr})}}\hspace{-1.5cm}{\vphantom{\mlpsum}}_{q_c(A)\ k_c}\prod\limits_{b=1}^{\mathrm{max}_{r}(A_{cr})} D_{bc}^{(1+n_{bc}(A))}(A)\notag
\\
&\ \hspace{2cm}\times  D_{q^{\{c\}}_{j_{\mathrm{max}_{r}(A_{cr})+1}^{\{c\}}+1}q^{\{c\}}_{j_{\mathrm{max}_{r}(A_{cr})+1}^{\{c\}}}}^{\left(\mathrm{max}_{r}(A_{cr})+1,2\right)}\mlpsum\limits_{({q_a^\prime}^{\{c\}})_{a=1}^{\mathrm{max}_{r}(A_{cr})+1} \vphantom{\begin{matrix}{}\\{}\\{}\\{}\end{matrix}}}\hspace{-0.5cm}{\vphantom{\mlpsum}}_{q_{j_{\mathrm{max}_{r}(A_{cr})+1}^{\{c\}}}^{\{c\}}k_{N+1}}\notag
\\
&\ \hspace{0.4cm}+ \mlpsum\limits_{(q_a^{\{c\}})_{a=1}^{j_c(A)} \vphantom{\begin{matrix}{}\\{}\end{matrix}}\ (j_b^{\{c\}})_{b=1}^{\mathrm{max}_{r}(A_{cr})}}\hspace{-1.5cm}{\vphantom{\mlpsum}}_{q_c(A)\ k_c}\sum\limits_{b=1}^{\mathrm{max}_{r}(A_{cr})} D_{bc}^{(2+n_{bc}(A))}(A)\notag
\\
&\ \hspace{2cm}\times \prod\limits_{\mbox{\scriptsize $ \begin{matrix}b^\prime=1\\ b^\prime\neq b \end{matrix} $ }}^{\mathrm{max}_{r}(A_{cr})} D_{b^\prime c}^{(1+n_{b^\prime c}(A))}\mlpsum\limits_{({q_a^{\prime\prime}}^{\{c\}})_{a=1}^{j_b^{\{c\}}} \vphantom{\begin{matrix}{}\\{}\\{}\\{}\end{matrix}}}\hspace{-0.2cm}{\vphantom{\mlpsum}}_{q_{j_{b}^{\{c\}}}^{\{c\}}k_{N+1}}\label{eq-commutator-2}
\end{align}
\vspace{-0.7cm}

\noindent
Due to the derivation, a new vertex is introduced to each summand. However, note that the way this new vertex is connected to the given vertices differs in both terms: In the first summand there now appears to be an additional propagator of second order establishing a connection to the $N+1^\text{th}$ vertex, while in the remaining ${\mathrm{max}}_r(A_{cr})$ summands, the order of the $b^\text{th}$ propagator is raised by one, which also allows an additional connection to the new vertex. Let us define the set
\begin{equation}
\nu_c(A) = \bigcup\limits_{b=1}^{\mathrm{max}_{r}(A_{cr})+1}\left\{ \begin{pmatrix} A_{11} & \ldots & A_{1N} & 0 \\ \vdots & \ddots & \vdots & \vdots \\ A_{c1} & \ldots & A_{cN} & b \\ \vdots & \ddots & \vdots & \vdots \\ A_{N1} & \ldots & A_{NN} & 0 \\
0 & 0 & 0 & 0  \end{pmatrix} \right\},
\end{equation}
which is a subset of $\mathbb{A}^{N+1}$. Its $|\nu_c(A)|=\mathrm{max}_{r}(A_{cr})+1$ elements correspond to adjacency matrices of N-vertex arborescences, that have been extended by an $(N+1)^\text{th}$ vertex, which is connected to the $c^\text{th}$ vertex. For the element with $b=\mathrm{max}_r(A_{cr})+1$, this corresponds to establishing a connection via an additional propagator, that is consequently of second order. Else, we have $b\in\{1,\ldots,\mathrm{max}_r(A_{cr})\}$, which corresponds to raising the order of the $b^\text{th}$ propagator in the $c^\text{th}$ vertex and thereby allows being connected with the new vertex.
\clearpage

The observations made above can be formulated in the language of adjacency matrices: In terms of $(N+1)\times (N+1)$ adjacency matrices of the set $\nu_c(A)$, the commutator in Eq.~\eqref{eq-commutator-2} is given by
\vfill
\noindent
\begin{align}
\left[ \frac{\partial}{\partial x_{k_{N+1}}}, B_c(A)\right] &= \Theta\left(l-\mathrm{max}_{r}(A_{cr})-1\right) B_c\left[ \begin{pmatrix} A_{11} & \ldots & A_{1N} & 0 \\ \vdots & \ddots & \vdots & \vdots \\ A_{c1} & \ldots & A_{cN} & \mathrm{max}_{r}(A_{cr})+1 \\ \vdots & \ddots & \vdots & \vdots \\ A_{N1} & \ldots & A_{NN} & 0 \\
0 & 0 & 0 & 0  \end{pmatrix}   \right] \notag\\
&\ \hspace{3.15cm}\times B_{N+1}\left[ \begin{pmatrix} A_{11} & \ldots & A_{1N} & 0 \\ \vdots & \ddots & \vdots & \vdots \\ A_{c1} & \ldots & A_{cN} & \mathrm{max}_{r}(A_{cr})+1 \\ \vdots & \ddots & \vdots & \vdots \\ A_{N1} & \ldots & A_{NN} & 0 \\
0 & 0 & 0 & 0  \end{pmatrix}   \right]\notag\\
&\ \hspace{0.2cm}+\sum\limits_{b=1}^{\mathrm{max}_{r}(A_{cr})}B_c\left[ \begin{pmatrix} A_{11} & \ldots & A_{1N} & 0 \\ \vdots & \ddots & \vdots & \vdots \\ A_{c1} & \ldots & A_{cN} & b \\ \vdots & \ddots & \vdots & \vdots \\ A_{N1} & \ldots & A_{NN} & 0 \\
0 & 0 & 0 & 0  \end{pmatrix}   \right] B_{N+1}\left[ \begin{pmatrix} A_{11} & \ldots & A_{1N} & 0 \\ \vdots & \ddots & \vdots & \vdots \\ A_{c1} & \ldots & A_{cN} & b \\ \vdots & \ddots & \vdots & \vdots \\ A_{N1} & \ldots & A_{NN} & 0 \\
0 & 0 & 0 & 0  \end{pmatrix}   \right].\label{eq-commutator-3}
\end{align}
\vfill
\noindent
Here we could express the $(N+1)^\text{th}$ vertex as a term $B_{N+1}(A^\prime)$ with $A^\prime\in\nu_c(A)$, due to the $(N+1)^\text{th}$ line containing only zeros, thus $\mathrm{max}_{r}({A^\prime}_{N+1,r})=0$, and due to $\beta_{N+1}(A^\prime)=c$ as well as ${A^\prime}_{\beta_{N+1}({A^\prime}),N+1}=b$,
\vfill
\noindent
\begin{align}
\mlpsum\limits_{({q_a^{\{N+1\}}})_{a=1}^{j_b^{\{c\}}} \vphantom{\begin{matrix}{}\\{}\\{}\\{}\end{matrix}}}\hspace{-0.2cm}{\vphantom{\mlpsum}}_{q_{j_{b}^{\{c\}}}^{\{c\}}k_{N+1}} &= \mlpsum\limits_{(q_a^{\{N+1\}})_{a=1}^{j_{N+1}(A^\prime)} \vphantom{\begin{matrix}{}\\{}\\{}\\{}\end{matrix}}\ (j_b^{\{N+1\}})_{b=1}^{\mathrm{max}_{r}({A^\prime}_{N+1,r})}}\hspace{-2.2cm}{\vphantom{\mlpsum}}_{q_{N+1}(A^\prime)k_{N+1}} =B_{N+1}\left[ \begin{pmatrix} A_{11} & \ldots & A_{1N} & 0 \\ \vdots & \ddots & \vdots & \vdots \\ A_{c1} & \ldots & A_{cN} & b \\ \vdots & \ddots & \vdots & \vdots \\ A_{N1} & \ldots & A_{NN} & 0 \\
0 & 0 & 0 & 0  \end{pmatrix}   \right].\notag
\end{align}
\vfill
\noindent
Each element of $\nu_c(A)$ is represented in this sum, which implies that each possible connection from the $c^\text{th}$ vertex to the $(N+1)^\text{th}$ vertex is established. Note that the first summand, that introduces a new propagator, is the only term that may alter the saturation threshold of the arborescence, namely in the case that $\mathrm{max}_r(A_{cr})\geq \alpha(A)$. Therefore, we write
\clearpage
\begin{equation}
 \Theta\left(l-\mathrm{max}_{r}(A_{cr})-1\right) \Theta(l-\alpha(A))= \Theta\left(l-\alpha\left[ \begin{pmatrix} A_{11} & \ldots & A_{1N} & 0 \\ \vdots & \ddots & \vdots & \vdots \\ A_{c1} & \ldots & A_{cN} & \mathrm{max}_{r}(A_{cr})+1 \\ \vdots & \ddots & \vdots & \vdots \\ A_{N1} & \ldots & A_{NN} & 0 \\
0 & 0 & 0 & 0  \end{pmatrix}   \right]\right)\label{eq-dominate-satthr}
\end{equation}
The derivative of the arborescence $\delta_{mk_1\ldots k_N}^{(l,N)}(A)$ can be written as
\begin{align}
\frac{\partial}{\partial x_{k_{N+1}}}\delta_{mk_1\ldots k_N}^{(l,N)}(A) &= \left[ \frac{\partial}{\partial x_{k_{N+1}}}, \delta_{mk_1\ldots k_N}^{(l,N)}(A)\right] \notag\\ &= \sum\limits_{q^{\{ 0\}}_l}\delta_{mq_l^{\{ 0\}}}\sum\limits_{j_{A_{01}}^{\{0\}}}\delta_{lj_{A_{01}}^{\{0\}}}\left[\frac{\partial}{\partial x_{k_{N+1}}},\prod\limits_{c=1}^N B_c(A)\right].\notag
\end{align}
Using the commutator relation in Eq.~\eqref{eq-commutator-1}, we can express $\delta_{mk_1\ldots k_N}^{(l,N)}(A)$ in terms of the individual commutators from Eq.~\eqref{eq-commutator-3},
\begin{equation}
\frac{\partial}{\partial x_{k_{N+1}}}\delta_{mk_1\ldots k_N}^{(l,N)}(A)= \sum\limits_{q^{\{ 0\}}_l}\delta_{mq_l^{\{ 0\}}}\sum\limits_{j_{A_{01}}^{\{0\}}}\delta_{lj_{A_{01}}^{\{0\}}}\sum\limits_{i=0}^{N-1} \left( \prod\limits_{j=1}^i B_j \right)\left[\frac{\partial}{\partial x_{k_{N+1}}},B_{i+1}\right]\left( \prod\limits_{j=i+2}^N B_j \right).\label{eq-proof-derivative}
\end{equation}
It is very insightful to analyze Eq.~\eqref{eq-proof-derivative}: We already know that the $i^\text{th}$ commutator $[\partial/\partial x_{k_{N+1}},B_{i}(A)]$ is a sum of $\mathrm{max}_r(A)+1$ terms and corresponds to establishing a connection from the $i^\text{th}$ vertex to the $(N+1)^\text{th}$ vertex, either by introducing a new propagator or by raising the order of an already existing propagator by one. However, Eq.~\eqref{eq-proof-derivative} is a sum of $N$ terms with the $i^\text{th}$ summand containing the $i^\text{th}$ commutator. This means that all allowed connections from all of the given $N$ vertices to the $(N+1)^\text{th}$ vertex are covered here. As the $i^\text{th}$ commutator leaves other vertices unaltered,
\begin{equation}
i \neq i^\prime \Rightarrow \forall A^\prime\in \nu_{i^\prime}(A): B_i(A^\prime) = B_i(A),\notag
\end{equation}
we can write for a single arborescence, using Eq.~\eqref{eq-dominate-satthr},
\begin{equation}
\Theta(l-\alpha(A))\frac{\partial}{\partial x_{k_{N+1}}}\delta_{mk_1\ldots k_N}^{(l,N)}(A) = \sum\limits_{A^\prime\in\nu(A)}\Theta(l-\alpha(A^\prime))\delta_{mk_1\ldots k_{N+1}}^{(l,N+1)}(A^\prime),\label{eq-proof-arbo-1}
\end{equation}
where we sum over the union
\begin{equation}
\nu(A) = \bigcup\limits_{c=1}^N \nu_c(A).\notag
\end{equation}
As the introduction of a new vertex to a given arborescence only influences the corresponding adjacency matrix by appending a new line and column, but leaves the original adjacency matrix unaltered, the disjuncture
\begin{equation}
A_1\neq A_2 \Rightarrow \nu(A_1)\cap\nu(A_2)=\emptyset\notag
\end{equation}
is obvious. Nonetheless, it can be easily argued that $\mathbb{A}^{N+1}$ is the union of all $\nu(A)$ for $A\in\mathbb{A}^N$. Therefore, it follows that both of the following sums must be identical:
\begin{equation}
\sum\limits_{A\in\mathbb{A}_N}\sum\limits_{A^\prime\in\nu(A)}\ldots = \sum\limits_{A\in\mathbb{A}_{N+1}}\ldots \notag
\end{equation}
Using Eq.~\eqref{eq-proof-arbo-1}, we finally complete the inductive step,
\begin{align}
\Delta_{mk_1\ldots k_{N+1}}^{(l,N+1)} &= \frac{\partial}{\partial x_{k_{N+1}}}\Delta_{mk_1\ldots k_N}^{(l,N)} \notag\\ 
&= \sum\limits_{A\in \mathbb{A}^N} \Theta(l-\alpha(A)) \frac{\partial}{\partial x_{k_{N+1}}} \delta_{mk_1\ldots k_N}^{(l,N)}(A)\notag\\
&= \sum\limits_{A\in \mathbb{A}^{N+1}} \Theta(l-\alpha(A)) \delta_{mk_1\ldots k_{N+1}}^{(l,N+1)}(A).\notag
\end{align}

\begin{thebibliography}{99}

\bibitem{Mehta:2014ms} 
	P.~Mehta and D.~J.~Schwab,
  ``An exact mapping between the variational renormalization group and deep learning,''
  arXiv:1410.3831 [stat.ML] (2014).
	
\bibitem{Baldi:2014bsw} 
  P.~Baldi, P.~Sadowski and D.~Whiteson,
  ``Searching for exotic particles in high-energy physics with deep learning,''
  Nature communications {\bf 5}.1, pp. 1-9 (2014).
	
\bibitem{Mills:2017} 
  K.~Mills, M.~Spanner and I.~Tamblyn,
  ``Deep learning and the Schrödinger equation,''
	Phys.~Rev.~A~{\bf 96}, 042113 (2017).
	
\bibitem{Richards:2011za} 
  J.~W.~Richards {\it et al.},
  ``On Machine-Learned Classification of Variable Stars with Sparse and Noisy Time-Series Data,''
  Astrophys.\ J.\  {\bf 733}, 10 (2011).
  

\bibitem{Buckley:2011kc} 
  A.~Buckley, A.~Shilton and M.~J.~White,
  ``Fast supersymmetry phenomenology at the Large Hadron Collider using machine learning techniques,''
  Comput.\ Phys.\ Commun.\  {\bf 183}, 960 (2012).

\bibitem{Graff:2013cla} 
  P.~Graff, F.~Feroz, M.~P.~Hobson and A.~N.~Lasenby,
  ``SKYNET: an efficient and robust neural network training tool for machine learning in astronomy,''
  Mon.\ Not.\ Roy.\ Astron.\ Soc.\  {\bf 441}, 1741 (2014).

\bibitem{Carleo:2017} 
  G.~Carleo and M.~Troyer,
  ``Solving the quantum many-body problem with artificial neural networks,''
	Science {\bf 355}, 602 (2017).
  
\bibitem{Wetzel:2017ooo} 
  S.~J.~Wetzel and M.~Scherzer,
  ``Machine Learning of Explicit Order Parameters: From the Ising Model to SU(2) Lattice Gauge Theory,''
  Phys.\ Rev.\ B {\bf 96},  184410 (2017).


\bibitem{He:2017set} 
  Y.~H.~He,
  ``Machine-learning the string landscape,''
  Phys.\ Lett.\ B {\bf 774}, 564 (2017).

  
\bibitem{Fujimoto:2017cdo} 
  Y.~Fujimoto, K.~Fukushima and K.~Murase,
  ``Methodology study of machine learning for the neutron star equation of state,''
  Phys.\ Rev.\ D {\bf 98}, 023019 (2018).

\bibitem{Wu:2018} 
  Y.~Wu, P.~Zhang, H.~Shen and H.~Zhai,
  ``Visualizing Neural Network Developing Perturbation Theory,''
  Phys.\ Rev.\ A {\bf 98}, 010701 (2018).

   
\bibitem{Niu:2018trk} 
  Z.~M.~Niu, H.~Z.~Liang, B.~H.~Sun, W.~H.~Long and Y.~F.~Niu,
  ``Predictions of nuclear $\beta$-decay half-lives with machine learning and their impact on r -process nucleosynthesis,''
  Phys.\ Rev.\ C {\bf 99},  064307 (2019).

\bibitem{Brehmer:2018kdj} 
  J.~Brehmer, K.~Cranmer, G.~Louppe and J.~Pavez,
  ``Constraining Effective Field Theories with Machine Learning,''
  Phys.\ Rev.\ Lett.\  {\bf 121},  111801 (2018).

\bibitem{Steinheimer:2019iso} 
  J.~Steinheimer, L.~Pang, K.~Zhou, V.~Koch, J.~Randrup and H.~Stoecker,
  ``A machine learning study to identify spinodal clumping in high energy nuclear collisions,''
  JHEP {\bf 1912}, 122 (2019).

  
\bibitem{Larkoski:2017jix} 
  A.~J.~Larkoski, I.~Moult and B.~Nachman,
  ``Jet Substructure at the Large Hadron Collider: A Review of Recent Advances in Theory and Machine Learning,''
  Phys.\ Rept.\  {\bf 841}, 1 (2020).
	
\bibitem{Cybenko:1989} 
  G.~Cybenko,
  ``Approximation by superpositions of a sigmoidal function,''
  Mathematics of control, signals and systems {\bf 2}.4, pp. 303-314 (1989).
	
\bibitem{Hornik:1991} 
  K.~Hornik,
  ``Approximation capabilities of multilayer feedforward networks,''
  Neural networks {\bf 4}.2, pp. 251-257 (1991).
	
\bibitem{Montavon:2017mlbsm} 
  G.~Montavon, S.~Lapuschkin, A.~Binder, W.~Samek and K.-R.~M\"uller,
  ``Explaining nonlinear classification decisions with deep Taylor decomposition,''
  Pattern Recognition {\bf 65}, pp. 211-222 (2017).

\bibitem{Ribeiro:2016rsg} 
  M.~T.~Ribeiro, S.~Singh, C.~Guestrin,
  ``" Why should I trust you?" Explaining the predictions of any classifier,''
	Proceedings of the {\bf 22}nd ACM SIGKDD International Conference on Knowledge Discovery and Data Mining, pp. 1135–1144 (2016).

\bibitem{Fan:2020fxw} 
	F.~Fan, J.~Xiong and G.~Wang,
  ``On interpretability of artificial neural networks,''
  arXiv:2001.02522 [cs.LG] (2020).

\bibitem{Hairer:1993hnw} 
  E.~Hairer, S.~P.~N{\o}rsett, G.~Wanner,
  Solving ordinary differential equations I. Nonstiff problems,
	Springer Berlin Heidelberg (1993).
	
\bibitem{Nielsen:2015} 
  M.~Nielsen,
  ``Neural networks and deep learning'',
	Determination Press (2015).
	
\bibitem{Kamiyama:2014k} 
  N.~Kamiyama,
  ``Arborescence problems in directed graphs: Theorems and algorithms,''
  Interdisciplinary information sciences {\bf 20}.1, pp. 51-70 (2014).
	
	\bibitem{Jonsson:1990} 
  B.~Jonsson, S.T.~Eng,
  ``Solving the Schrodinger equation in arbitrary quantum-well potential profiles using the transfer matrix method,''
	IEEE journal of quantum electronics {\bf 26}.11, pp. 2025-2035 (1990).
	
\bibitem{Paszke:2019} 
  A.~Paszke, S.~Gross,  F.~Massa, A.~Lerer, J.~Bradbury,  G.~Chanan, T.~Killeen, Z.~Lin, N.~Gimelstein,
  L.~Antiga, A.~Desmaison, A.~Kopf, E.~Yang, Z.~DeVito, M.~Raison, T.~Alykhan, S.~Chilamkurthy,
  B.~Steiner, F.~Lu, J.~Bai and   S.~Chintala,
  PyTorch: An imperative style, high-performancedeep learning library, 
  in: Advances in Neural Information Processing Systems {\bf 32},
  pp. 8024-8035 (eds. H.~Wallach, H.~Larochelle, A.~Beygelzimer, F.~d'Alch\'e-Buc, E.~Fox and R.~Garnett) (2019).
	
\bibitem{Hendrycks:2020} 
  D.~Hendrycks, K.~Gimpel,
  ``Barron, Jonathan T. "Gaussian error linear units (GELUs),''
  arXiv:1606.08415v4 [cs.LG] (2020).
	
\bibitem{Liu:2019lzgql} 
  Y.~Liu, J.~Zhang, C.~Gao, J.~Qu and L.~Ji,
  ``Natural-Logarithm-Rectified Activation Function in Convolutional Neural Networks,''
  5th International Conference on Computer and Communications (ICCC), pp. 2000-2008 (2019).
	
\bibitem{He:2015hzrs} 
  K.~He, X.~Zhang, S.~Ren and J.~Sun,
  ``Delving deep into rectifiers: Surpassing human-level performance on imagenet classification,''
  IEEE Proceedings, pp. 1026-1034 (2015).
	
	
\bibitem{Kingma:2017kb} 
  D.~P.~Kingma and J.~L.~Ba,
  ``Adam: A method for stochastic optimization,''
  arXiv:1412.6980v9 [cs.LG] (2017).
	
\bibitem{Loshchilov:2019lh} 
  I.~Loshchilov and F.~Hutter,
  ``Decoupled weight decay regularization,''
  arXiv:1711.05101v3 [cs.LG] (2019).
	
\end{thebibliography}
\end{document}